\documentclass{article}

\usepackage{microtype}
\usepackage{graphicx}
\usepackage{subcaption}
\usepackage{booktabs} 
\usepackage{tabularx} 
\usepackage{hyperref}

\usepackage[accepted]{icml2026}

\usepackage{amsmath}
\usepackage{amssymb}
\usepackage{mathtools}
\usepackage{amsthm}\usepackage[utf8]{inputenc} 
\usepackage[T1]{fontenc}    
\usepackage{hyperref}       
\usepackage{url}            
\usepackage{booktabs}       
\usepackage{amsfonts}       
\usepackage{nicefrac}       
\usepackage{microtype}      
\usepackage{xcolor}         
\usepackage{amsmath}
\usepackage{amssymb}
\usepackage{algorithm}
\usepackage{amsthm}
\usepackage[english]{babel}
\usepackage{graphicx}
\usepackage{bm}
\usepackage{wrapfig}
\usepackage{caption}
\usepackage{subcaption}
\usepackage{booktabs}
\usepackage{tikz}
\usepackage{pifont} 
\usepackage{multirow} 

\newcommand{\cmark}{\ding{51}} 
\newcommand{\xmark}{\ding{55}} 

\renewcommand{\algorithmicrequire}{\textbf{Inputs:}}
\renewcommand{\algorithmicensure}{\textbf{Output:}}

\newcommand{\bb}[1]{\bm{\mathrm{#1}}}
\newcommand{\CommentAbove}[1]{\Statex\hspace{\algorithmicindent}\hspace{\algorithmicindent}\textit{#1}}
\def\RR{\mathbb{R}}

\def\Bcal{\mathcal{B}}
\def\Ccal{\mathcal{C}}
\def\Dcal{\mathcal{D}}
\def\Ecal{\mathcal{E}}
\def\Fcal{\mathcal{F}}
\def\Gcal{\mathcal{G}}
\def\Lcal{\mathcal{L}}
\def\Pcal{\mathcal{P}}
\def\Rcal{\mathcal{R}}
\def\Scal{\mathcal{S}}

\usepackage[capitalize,noabbrev]{cleveref}

\theoremstyle{plain}
\newtheorem{theorem}{Theorem}[section]

\newtheorem{lemma}[theorem]{Lemma}

\theoremstyle{definition}

\theoremstyle{remark}

\usepackage[disable,textsize=tiny]{todonotes}

\icmltitlerunning{L-SR1: Learned Symmetric-Rank-One Preconditioning}

\begin{document}

\twocolumn[
  \icmltitle{L-SR1: Learned Symmetric-Rank-One Preconditioning}



  \icmlsetsymbol{equal}{*}

  \begin{icmlauthorlist}
    \icmlauthor{Gal Lifshitz}{tau}
    \icmlauthor{Shahar Zuler}{tau}
    \icmlauthor{Ori Fouks}{tau}
    \icmlauthor{Dan Raviv}{tau}
  \end{icmlauthorlist}

  \icmlaffiliation{tau}{Tel Aviv University, Tel Aviv, Israel}
  \icmlcorrespondingauthor{Gal Lifshitz}{lifshitz@mail.tau.ac.il}

  \icmlkeywords{Machine Learning, ICML}

  \vskip 0.3in
]



\printAffiliationsAndNotice{}  

 \begin{abstract}
End-to-end deep learning has achieved impressive results but often relies on large labeled datasets, exhibits limited generalization to unseen scenarios, and incurs substantial computational cost. 
Classical optimization methods, in contrast, are more data-efficient and lightweight but frequently suffer from slow convergence. 
Learned optimizers aim to bridge this gap, yet existing approaches have focused primarily on first-order methods, while learned second-order optimization has received much less attention. 
We introduce L-SR1, a learned second-order optimizer inspired by the classical Symmetric Rank-One (SR1) method. 
At its core, L-SR1 employs a Projection-Guided Secant Mechanism (PGSM) that generates positive semi-definite preconditioners and biases meta-training toward the quasi-Newton secant relation. 
Through controlled analytic benchmarks, we study stability, generalization across problem dimensions, and search direction quality, and further evaluate L-SR1 on Monocular Human Mesh Recovery (HMR), where it outperforms both classical and learned optimization-based baselines. 
With a compact model and no reliance on task-specific fine-tuning or annotated data, L-SR1 demonstrates strong generalization and can be integrated into a broad range of iterative optimization problems to accelerate convergence and reduce the required number of iterations.

\end{abstract}

\section{Introduction}
\begin{figure*}[t]
     \centering
     \begin{subfigure}[b]{0.32\textwidth}
         \centering
         \includegraphics[width=\textwidth]{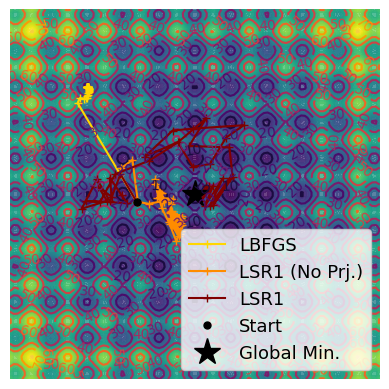}  
         \caption{\textbf{Rastrigin function.}}
         \label{fig:trj_rast}
     \end{subfigure}
     \hfill
     \begin{subfigure}[b]{0.32\textwidth}
         \centering
         \includegraphics[width=\textwidth]{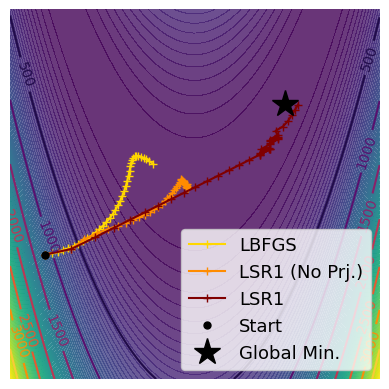}
         \caption{\textbf{Rosenbrock function.}}
         \label{fig:trj_rosen}
     \end{subfigure}
     \hfill
     \begin{subfigure}[b]{0.32\textwidth}
         \centering
         \includegraphics[width=\textwidth]{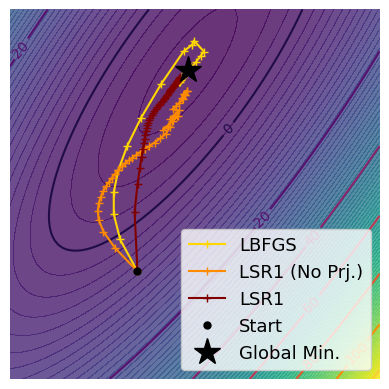}
         \caption{\textbf{Quadratic function.}}
         \label{fig:trj_quad}
     \end{subfigure}
    \caption{\textbf{Optimization trajectories.}  
    Example trajectories of LBFGS, L-SR1 with PGSM, and L-SR1 without (No Prj.) on a quadratic function and two well-known challenging benchmark functions \citep{simulationlib}---the Rosenbrock and Rastrigin functions. PGSM improves the performance of L-SR1 compared to the variant without PGSM (No Prj.), while maintaining model compactness.}
        \label{fig:trjs}
\end{figure*}

End-to-end deep learning has demonstrated significant power but is constrained by its reliance on large labeled datasets and limited ability to generalize to unseen scenarios. Furthermore, increased model sizes featured in recent works pose a limitation as they demand high compute and memory resources.
In contrast, classical optimization excels in data-scarce settings and features a low memory stamp, but often suffers from long runtime due to its iterative nature. To address this, extensive research has focused on accelerating convergence, with optimization methods broadly categorized into first-order and second-order approaches.

First-order methods, such as Adam \citep{kingma2014adam} and Nesterov Accelerated Gradient (NAG) \citep{nesterov1983method,sutskever2013importance}, rely on estimated gradient momentums for parameter updates. Second-order methods, such as Symmetric-Rank-One (SR1) \citep{conn1991convergence} and Broyden-Fletcher-Goldfarb-Shanno (BFGS) \citep{liu1989limited}, utilize approximations of the inverse Hessian matrix \citep{boyd2004convex,Bertsekas/99}. While more computationally intensive, second-order methods typically achieve faster convergence by accurately capturing the underlying structure of the objective function and exploiting dependencies between variables.

Learned optimization has recently emerged as a promising field that leverages deep learning to enhance traditional optimization methods. These approaches incorporate trainable deep neural network (DNN) architectures—such as Multi-Layer Perceptrons (MLPs) \citep{andrychowicz2016learning,li2016learning,song2020human}, Recurrent Neural Networks (RNNs) \citep{andrychowicz2016learning}, Transformers \citep{gartner2023transformer}, and hybrid models \citep{metz2020tasks}—into optimization frameworks. Once trained on specific objectives, these learned optimizers exhibit significantly faster convergence. 
Although learned optimization has been widely explored for first-order methods~\citep{metz2022practical}, several works have also considered second-order approaches~\citep{li2020learning,liao2023learningoptimizequasinewtonmethods}. Nevertheless, the integration of learnable components with second-order methods remains relatively underexplored.

By accelerating convergence, learned optimizers offer a compelling path to bridging classical optimization techniques with modern deep learning-based approaches in computer vision tasks.

Monocular Human Mesh Recovery (HMR) seeks to estimate a 3D human mesh from a single 2D image—a fundamentally ill-posed problem due to the inherent loss of depth information. Historically, in the absence of large-scale annotated datasets, optimization-based methods dominated the field \citep{bogo2016keep,pavlakos2019expressive}. However, with the emergence of increasingly comprehensive annotated datasets, these traditional approaches were largely supplanted by deep neural network (DNN)-based methods. These newer methods span both end-to-end regression models and iterative frameworks \citep{sun2023trace,shin2024wham,wang2025prompthmr}, including learned optimization \citep{kolotouros2019learning,song2020human}, achieving state-of-the-art performance. Nonetheless, they come with notable trade-offs, including significantly larger model sizes and a continued reliance on vast amounts of annotated training data.

In this work, we introduce Learned-SR1 (L-SR1), a novel learned second-order optimizer that extends the classical SR1 algorithm. Our method incorporates a learnable module that generates data-driven rank-one updates to approximate the inverse Hessian, enabling more informed and efficient steps. 
To preserve core quasi-Newton properties, we propose a \emph{Projection-Guided Secant Mechanism} (PGSM) that parameterizes the preconditioner through learned rank-one outer products, yielding positive semi-definite matrices by construction. During meta-training, PGSM penalizes secant mismatch in the spirit of least-squares projection onto PSD matrices, without requiring a projection subproblem at inference (Section~\ref{sec:projection}).
L-SR1 instantiates PGSM with a limited-memory buffer of these vectors and lightweight trainable modules. Building on the SR1 framework, it achieves state-of-the-art performance with a notably compact model size.

We evaluate our approach on both analytic benchmark functions (Fig.~\ref{fig:trjs}) and the real-world task of HMR. Through controlled analytic experiments, we study generalization across problem dimensions and the quality of the search directions produced by L-SR1, providing insight into its behavior beyond a single downstream application.
On HMR, a challenging high-dimensional, non-convex optimization problem in which each iteration involves a computationally intensive 3D-to-2D projection, L-SR1 converges efficiently and integrates robustly into the iterative optimization pipeline without requiring task-specific fine-tuning.
Together, these results highlight the potential of L-SR1 as a general optimization framework that can be incorporated into a wide range of optimization-based pipelines.

We summarize our key contributions as follows:
\begin{itemize}
    \item We propose Learned-SR1 (L-SR1), a lightweight, self-supervised learned optimizer that integrates a trainable preconditioning unit into the SR1 framework, enabling data-driven curvature estimation without the need for annotated data or supervised meta-training.
    \item We introduce PGSM, a projection-guided secant mechanism with a PSD-by-construction rank-one parameterization and a meta-training secant-violation penalty motivated by PSD projection, preserving core quasi-Newton structure without extra inference cost.
    \item We demonstrate that L-SR1 can be integrated into optimization-based HMR pipelines, where it consistently outperforms classical and learned solvers in terms of convergence and generalization.
\end{itemize}

\paragraph{Conflict of Interest Disclosure.}
The authors have no financial conflicts of interest to disclose regarding this work.

\section{Related Work}
\paragraph{Second-Order Optimization and Preconditioning}
Second-order optimization methods leverage curvature information to accelerate convergence, typically by using the inverse Hessian matrix as a preconditioner \citep{boyd2004convex,Bertsekas/99}. Since computing and inverting the full Hessian is often impractical, Quasi-Newton methods such as DFP and BFGS \citep{Bertsekas/99}, LBFGS \citep{nocedal1980updating, liu1989limited}, and SR1 \citep{conn1991convergence, khalfan1993theoretical} were introduced to iteratively estimate this preconditioner. SR1, in particular, uses rank-one updates and demonstrated favorable convergence under mild conditions. These ideas were extended to large-scale problems in deep learning using scalable approximations \citep{martens2015optimizing, gupta2018shampoo, yao2021adahessian}.

\paragraph{Learned Optimization and Model-Based Deep Learning}
Gradient-based methods are lightweight and broadly applicable but do not explicitly model curvature information. Classical Quasi-Newton methods, in contrast, exploit curvature structure at the cost of more expensive updates. Learned optimization seeks to bridge classical optimization principles with the adaptability of deep learning by embedding trainable components into iterative solvers.

Learned optimization methods generate adaptive update rules from data \citep{andrychowicz2016learning, li2016learning, metz2020tasks, wichrowska2017learned, metz2022practical, metz2022velo, chen2022learning}. While early works primarily focused on first-order optimization dynamics, more recent approaches introduced richer architectures, including Transformer-based optimizers \citep{gartner2023transformer} and low-rank attention mechanisms \citep{jain2023mnemosyne}. In parallel, model-based deep learning integrated learning into principled algorithmic formulations to preserve interpretability and robustness \citep{shlezinger2020viterbinet, revach2022kalmannet, shlezinger2023model}.

Several works have explored learned optimizers inspired by second-order methods and Quasi-Newton optimization. \citep{li2020learning} proposed a learned optimizer motivated by Quasi-Newton principles, while \citep{ayad2024qn} developed an unrolled BFGS-like network for CT reconstruction. In addition, \citep{gartner2023transformer} introduced a general-purpose learned second-order optimizer, and \citep{liao2023learningoptimizequasinewtonmethods} learned a preconditioning matrix that is updated online during optimization without explicit meta-training.

In contrast to previous work, our approach introduces PGSM---a projection-guided secant mechanism with a PSD-by-construction rank-one parameterization and a meta-training secant penalty---within a lightweight, self-supervised SR1-inspired preconditioner that stores the learned rank-one vectors in a limited-memory buffer at inference.

\paragraph{Human Mesh Recovery (HMR)}
Human Mesh Recovery (HMR) aims to estimate 3D human meshes from single RGB images, a highly ill-posed problem due to the loss of depth. Early works approached HMR through optimization-based fitting \citep{bogo2016keep, pavlakos2019expressive}, which, while data-efficient, were slow and sensitive to initialization. With the emergence of large-scale datasets, deep learning methods were introduced, including regression-based approaches \citep{shin2024wham, sun2023trace,wang2025prompthmr} and iterative refinement techniques, including learned optimization \citep{kolotouros2019learning,song2020human,shetty2023pliks}. These models achieved impressive performance but depended heavily on annotated data and large architectures. In contrast, our method integrates a learned SR1-inspired optimizer into the HMR process, outperforming learned
optimization-based methods while requiring neither large models nor explicit fine-tuning.

\section{Theoretical Background and Preliminaries} \label{sec:theory}

\subsection{Learned Optimization}

A typical learned optimization framework consists of the following update rule:
\begin{equation}
    \bb{x}_{k+1} \gets \bb{x}_k + \varphi_{\Theta}\left(\nabla f(\bb{x}_k), \bb{x}_k, \ldots \right),
\end{equation}
where $\varphi_{\Theta}(\cdot)$ is a learnable function parameterized by $\Theta$, which can be conditioned on a variety of features, such as the current iterate and gradient.

Training a learned optimizer (\emph{meta-training}) alternates between \emph{inner} and \emph{outer} iterations. In each outer iteration, the optimizer performs $K$ unrolled inner steps, whose objective values are used to compute a \emph{meta-loss}, often of the form:
\begin{equation}
    \Lcal_{\text{meta}} = \sum_{k=1}^K f(\bb{x}_k),
\end{equation}
with optional additional supervision terms to leverage task-specific signals.

Gradients of this meta-loss with respect to $\Theta$ are then backpropagated through the unrolled computation graph, and the parameters are updated using a \emph{meta-optimizer}.

\subsection{Quasi-Newton Methods}

Let $f:\RR^n \rightarrow \RR$ be a twice differentiable objective function. The Quasi-Newton (QN) update step is given by
\begin{equation} \label{eq:qn}
    \bb{x}_{k+1} \gets \bb{x}_k - \alpha_k \bb{B}_k \bb{g}_k,
\end{equation}
where $\bb{B}_k$ is a preconditioning matrix approximating the inverse Hessian at $\bb{x}_k$, $\bb{g}_k = \nabla f(\bb{x}_k)$ is the gradient, and $\alpha_k$ is a step size.

QN methods differ in how they update $\bb{B}_k$, but they all satisfy the \emph{secant constraint}, derived from a first-order Taylor approximation. Defining $\bb{p}_k = \bb{x}_{k} - \bb{x}_{k-1}$ and $\bb{q}_k = \bb{g}_k - \bb{g}_{k-1}$, the secant condition is
\begin{equation} \label{eq:sec}
    \bb{B}_k \bb{q}_k = \bb{p}_k.
\end{equation}

This condition ensures that the preconditioner $\bb{B}_k$ captures local curvature information. Additionally, $\bb{B}_k$ is typically required to be positive semi-definite to guarantee that the update direction is a descent direction.
The standard descent argument associated with Eq.~(\ref{eq:qn}) assumes a symmetric positive semi-definite $\bb{B}_k$ and a scalar step size $\alpha_k>0$.

In L-SR1 (Section~\ref{sec:lsr1}, Alg.~\ref{alg:lsr1}), the matrix is symmetric positive semi-definite by construction, but the trainable update applies a coordinate-wise vector of learning rates $\boldsymbol{\alpha}_k$ to the preconditioned direction.
Equivalently, the map from $\bb{g}_k$ to the step contains a diagonal left factor, so the composite linear operator need not be symmetric and the PSD-based intuition alone does not automatically yield a descent guarantee on the full update; practical behavior is therefore established empirically through meta-training and the experiments in Section~\ref{sec:experiments}.

\section{Learned Symmetric-Rank-One (L-SR1)} \label{sec:lsr1}
\begin{figure*}[t]
\centering
\includegraphics[width=\textwidth]{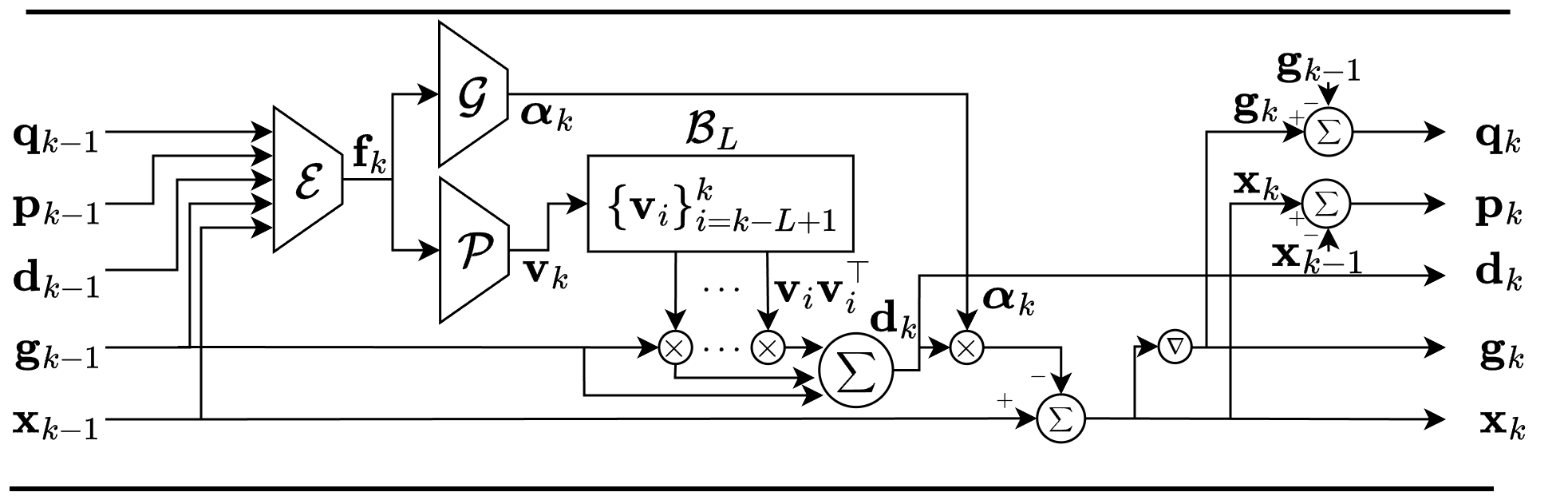}
\caption{
\textbf{Learned-SR1 (L-SR1) iteration block diagram.} At each iteration \(k\), the Input Encoder \(\mathcal{E}\) receives the vectors $\mathbf{x}_{k-1}$, $\mathbf{p}_{k-1}$, $\mathbf{d}_{k-1}$, $\mathbf{g}_{k-1}$, and $\mathbf{q}_{k-1}$, producing a feature vector \(\mathbf{f}_k\). This is passed to the Vector Generator \(\mathcal{P}\), which outputs a new direction vector \(\mathbf{v}_k\), and to the Learning Rate Generator \(\mathcal{G}\), which produces element-wise learning rates \(\boldsymbol{\alpha}_k\) (Sec. \ref{sec:componenets})). The updated descent direction \(\mathbf{d}_k\) is computed as a sum of rank-one terms \(\mathbf{v}_i \mathbf{v}_i^\top \mathbf{g}_{k-1}\), using the last \(L\) vectors stored in the buffer \(\mathcal{B}_L\). Finally, the optimization step is performed using \(\boldsymbol{\alpha}_k\) and \(\mathbf{d}_k\).
}
\label{fig:lsr1}
\end{figure*}

\begin{algorithm}[t]
\small
\caption{Learned-SR1 (L-SR1)}\label{alg:lsr1}
\begin{algorithmic}
\REQUIRE Objective $f \in \Ccal^2$, initial point $\bb{x}_0 \in \RR^n$, initial gradient $\bb{g}_0$, buffer $\Bcal_L = \{\emptyset\}$
\STATE Initialize $\bb{p}_0 \gets \bb{x}_0$, $\bb{q}_0 \gets \bb{g}_0$, $\bb{d}_0 \gets \bb{g}_0$
\FOR{$k = 1,2,\ldots$} 
    \STATE $\bb{f}_k \gets \Ecal(\bb{x}_{k-1}, \bb{p}_{k-1}, \bb{d}_{k-1}, \bb{g}_{k-1}, \bb{q}_{k-1})$
    \STATE $\bb{v}_k \gets \Pcal(\bb{f}_k)$,
           $\boldsymbol{\alpha}_k \gets \Gcal(\bb{f}_k)$
    \IF{$|\Bcal_L| = L$} 
        \STATE Discard oldest element in $\Bcal_L$
    \ENDIF
    \STATE $\Bcal_L \gets \Bcal_L \cup \bb{v}_k$
    \STATE $\bb{d}_k \gets \bb{g}_{k-1} + \sum_{\bb{v} \in \Bcal_L} \bb{v} \bb{v}^\top \bb{g}_{k-1}$
    \STATE $\bb{x}_k \gets \bb{x}_{k-1} - \boldsymbol{\alpha}_k \odot \bb{d}_k$
    \STATE $\bb{g}_k \gets \nabla f(\bb{x}_k)$
    \STATE $\bb{p}_k \gets \bb{x}_k - \bb{x}_{k-1}$,
           $\bb{q}_k \gets \bb{g}_k - \bb{g}_{k-1}$
\ENDFOR
\ENSURE $\mathbf{x}^* \gets \mathbf{x}_k$
\end{algorithmic}
\end{algorithm}

L-SR1 is a learned extension of the classical Symmetric Rank-One (SR1) Quasi-Newton method, designed to integrate lightweight trainable modules into a principled second-order optimization framework. The method enhances convergence while maintaining scalability and generalization across problem dimensions.

At its core, SR1 approximates the inverse Hessian matrix using a rank-one update of the form 
\begin{equation} \label{eq:sr1}
    \mathbf{B}_{k} \gets \mathbf{B}_{k-1} + \mathbf{u}_k \mathbf{v}_k^\top,
\end{equation}
where the vectors $\mathbf{v}_k = \mathbf{p}_k - \mathbf{B}_{k-1} \mathbf{q}_k$ and 
$\mathbf{u}_k = \mathbf{v}_k / (\mathbf{v}_k^\top \mathbf{q}_k)$ 
are chosen to satisfy the secant constraint presented in Eq. (\ref{eq:sec}).
This ensures that the update captures local curvature information of the objective function.

A significant advantage of the SR1 structure is that it relies solely on outer products of low-dimensional vectors. 
This allows for a limited-memory implementation, where instead of storing the full matrix $\mathbf{B}_k$, 
L-SR1 maintains a fixed-size buffer $\mathcal{B}_L$ of capacity $L$ containing the most recent $L$ vectors. 
These vectors are used to reconstruct $\mathbf{B}_k$ implicitly during optimization. 
If the buffer exceeds its capacity, the oldest entries are discarded, ensuring memory efficiency in high-dimensional problems.

A central design principle in L-SR1 is invariance to the problem dimension. 
All learnable components are constructed to operate element-wise, 
ensuring that the optimizer generalizes across problem sizes without re-training.

However, a known limitation of SR1 is that its updates do not guarantee positive semi-definiteness of $\mathbf{B}_k$, 
which can result in non-descent directions and instability. 
In the learned optimization realm, a naive fix is to use outer products of the form $\mathbf{v}\mathbf{v}^\top$ as was done in \citep{gartner2023transformer}, 
which are always positive semi-definite and symmetric, but such updates fail to satisfy the secant constraint. 
Traditional methods like LBFGS address this using more elaborate update strategies. 
In contrast, L-SR1 addresses this limitation through PGSM (Section~\ref{sec:projection}): a PSD-by-construction rank-one parameterization together with a meta-training secant penalty motivated by projection, without solving an explicit projection at each inner step.

A summary of the proposed method is given in Alg.~\ref{alg:lsr1} and a block diagram is in Fig~\ref{fig:lsr1}. We now describe the components of the L-SR1 algorithm in detail. 
The trainable modules are introduced in Section~\ref{sec:componenets}; PGSM is defined in Section~\ref{sec:projection}.

\subsection{Learned Components} \label{sec:componenets}
To enrich SR1 with data-driven flexibility, L-SR1 integrates three neural modules following the model-based deep learning paradigm of embedding learned operations inside a classical update while preserving structure~\citep{shlezinger2023model}.
All modules follow a standard and lightweight multi-layer perceptron (MLP) architecture and operate elementwise over the input dimensions, ensuring compatibility with varying problem sizes~\citep{metz2022practical,metz2022velo,chen2022learning}.
Model architectures are presented in Appendix~\ref{ap:lsr1_arch}.

\paragraph{Input Encoder $\mathcal{E}$}
The encoder constructs a latent representation of the optimization state. It takes as input 
the current point \(\mathbf{x}_{k-1}\), step \(\mathbf{p}_{k-1}\), descent direction \(\mathbf{d}_{k-1}\), gradient \(\mathbf{g}_{k-1}\), and gradient step \(\mathbf{q}_{k-1}\),
which are concatenated into an array of shape $N \times 5$, where $N$ is the problem dimension. 
The encoder then maps this input to a latent representation $\bb{f}_k \in \RR^{N \times M}$, with $M > 5$, lifting the per-coordinate features into a richer representation---as in prior per-parameter learned optimizers~\citep{gartner2023transformer,metz2022practical,metz2022velo}---that informs the subsequent modules.

\paragraph{Vector Generator $\mathcal{P}$}
This module produces a single vector $\mathbf{v}_k \in \mathbb{R}^N$ given $\bb{f}_k$ at each iteration. Under PGSM (Section~\ref{sec:projection}), outer products $\mathbf{v}_k \mathbf{v}_k^\top$ build a PSD preconditioner $\tilde{\mathbf{B}}_k$ and the secant penalty guides meta-training; a limited-memory buffer stores the most recent $L$ vectors for efficient inference.

\paragraph{Learning Rate Generator $\mathcal{G}$}
The learning rate generator takes the latent representation from the encoder $\bb{f}_k$ and outputs a vector $\tilde{\boldsymbol{\alpha}}_k \in \mathbb{R}^N$, interpreted as the logarithm of coordinate-wise learning rates, following~\citep{gartner2023transformer}. The final learning rate vector $\boldsymbol{\alpha}_k$ is computed element-wise using the transformation:
\[
\boldsymbol{\alpha}_k = \gamma_1 \cdot \exp\left( \gamma_2 \cdot \tilde{\boldsymbol{\alpha}}_k \right),
\]
where $\gamma_1$ and $\gamma_2$ are fixed scalar hyperparameters (not learned). Appendix~\ref{sec:lr_param} compares coordinate-wise and scalar learning-rate parameterizations, supporting this design choice.

\subsection{Projection-Guided Secant Mechanism (PGSM)} \label{sec:projection}

\begin{figure*}[t]
     \centering
     \begin{subfigure}[b]{0.32\textwidth}
         \centering
         \includegraphics[width=\textwidth]{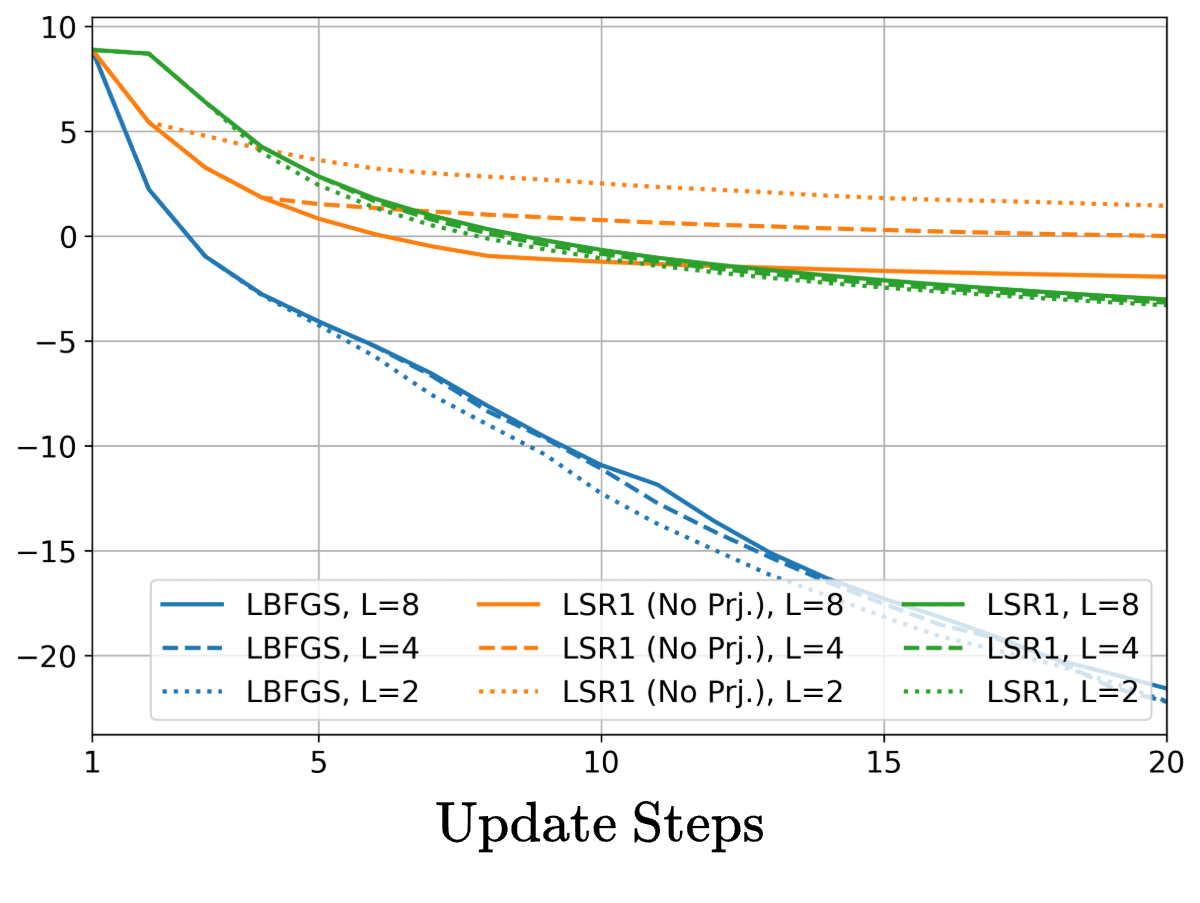}
         \caption{\textbf{Quadratic objective values.}}
         \label{fig:vals}
     \end{subfigure}
     \hfill
     \begin{subfigure}[b]{0.32\textwidth}
         \centering
         \includegraphics[width=\textwidth]{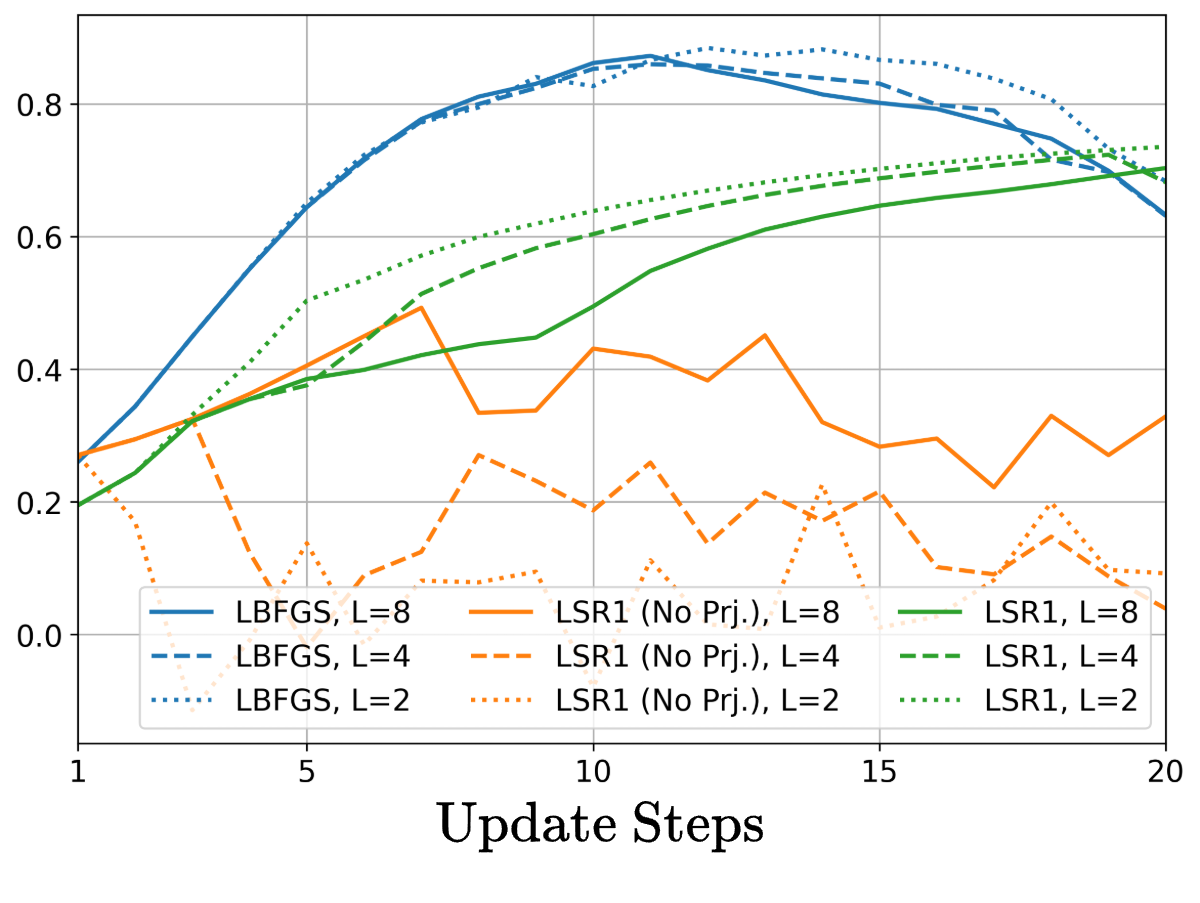}  
         \caption{\textbf{Cosine similarity to NM.}}
         \label{fig:dirs}
     \end{subfigure}
     \hfill     
     \begin{subfigure}[b]{0.32\textwidth}
         \centering
         \includegraphics[width=\textwidth]{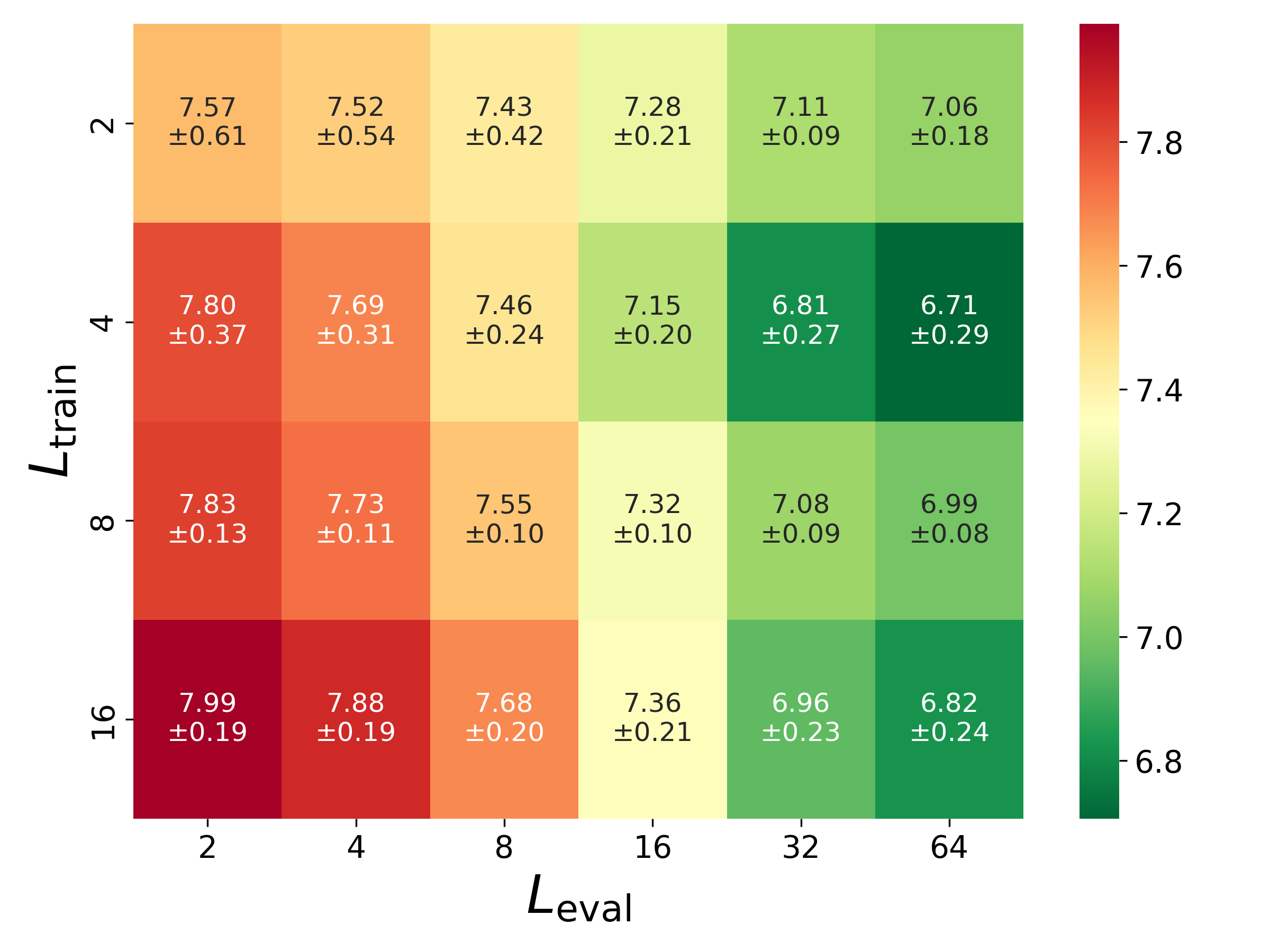}
         \caption{\textbf{AUC across buffer sizes.}}
         \label{fig:buffers}
     \end{subfigure}
\caption{\textbf{Quadratic functions experiments.} 
Figures~\ref{fig:vals} and~\ref{fig:dirs} show results from our quadratic experiments (Section~\ref{sec:quadratic}), comparing L-SR1 with PGSM and without (No Prj.) to LBFGS~\citep{nocedal1980updating}, which serves as a reference. 
Figures~\ref{fig:vals} and~\ref{fig:dirs} report objective values and cosine similarities with the Newton direction, respectively, averaged over the test set. 
PGSM improves L-SR1 over the variant without PGSM (No Prj.) in both convergence and alignment with the Newton direction.
Figure~\ref{fig:buffers} shows the area under loss curve (AUC) as a function of evaluation buffer size for models trained with different training buffer sizes, averaged over different seeds. Performance improves monotonically with evaluation buffer, and remains robust across training buffers.
}
\label{fig:analytic}
\end{figure*}

As discussed earlier, our goal is to construct preconditioning matrices $\mathbf{B}_k$ that are both positive semi-definite (PSD) and approximately satisfy the secant equation (Eq.~(\ref{eq:sec})). We motivate this objective through a projection formulation that minimizes the violation of the secant condition:
\begin{equation}
    \mathbf{B}_k^* = \pi_+(\mathbf{B}_k) = \underset{\mathbf{B}_k \in S_+}{\text{argmin}} \| \mathbf{p}_k - \mathbf{B}_k \mathbf{q}_k \|_2^2,
\end{equation}
where $S_+$ denotes the space of positive semi-definite matrices. This formulation seeks a PSD matrix that best approximates the secant constraint in a least-squares sense and motivates the proposed Projection-Guided Secant Mechanism (PGSM).

To ensure scalability and efficient computation, we constrain $\mathbf{B}_k$ to take the following structured form:
\begin{equation}
    \tilde{\mathbf{B}}_k = \mathbf{B}_0 + \sum_{i=1}^L \mathbf{v}_i \mathbf{v}_i^\top,
\end{equation}
where each vector $\mathbf{v}_i \in \mathbb{R}^n$ is produced by the vector generator module $\mathcal{P}$ and stored in a fixed-size buffer of length $L$. This construction ensures that $\tilde{\mathbf{B}}_k$ remains positive semi-definite as long as $\mathbf{B}_0 \succcurlyeq \bb{0}$, which we initialize as the identity matrix. The use of outer products is inspired by the classical SR1 update structure and is also consistent with techniques adopted in prior learned optimization works, such as \citep{gartner2023transformer}, which promote symmetry and positivity.

Within PGSM, the projection objective is formulated as the following optimization problem:
\begin{equation} \label{eq:our_proj}
    \tilde{\mathbf{B}}_k^* = \tilde{\pi}_+(\tilde{\mathbf{B}}_k) = \underset{\tilde{\mathbf{B}}_k = \mathbf{B}_0 + \sum_{i=1}^L \mathbf{v}_i \mathbf{v}_i^\top}{\text{argmin}} \| \mathbf{p}_k - \tilde{\mathbf{B}}_k \mathbf{q}_k \|_2^2.
\end{equation}

Rather than solving (\ref{eq:our_proj}) explicitly, PGSM uses the corresponding mismatch objective as the \emph{secant penalty} $\mathcal{R}_{\text{sec}}$, which is incorporated into the overall meta-training loss:
\begin{equation} \label{eq:loss}
    \mathcal{L}_{\text{meta}} = \frac{1}{K} \sum_{k=1}^K \left( f(\mathbf{x}_k) + \lambda_{\text{sec}} \cdot \| \mathbf{p}_k - \tilde{\mathbf{B}}_k \mathbf{q}_k \|_2^2 \right),
\end{equation}
where $\lambda_{\text{sec}}$ is a hyperparameter controlling the importance of satisfying the secant condition, and $K$ is the number of unrolled optimization steps per meta-iteration.

PGSM therefore maintains PSD preconditioning matrices by construction while encouraging approximate secant consistency through meta-training; no projection subproblem is solved at inference. L-SR1 applies the resulting preconditioner by reconstructing $\tilde{\mathbf{B}}_k$ from the limited-memory vector buffer, with no additional overhead beyond rank-one accumulation.

\section{Experimentation} \label{sec:experiments}

We conduct a series of experiments to evaluate both the performance and generalization properties of the proposed L-SR1 optimizer. We begin with controlled analytic experiments on quadratic objectives (Section~\ref{sec:quadratic}), where we isolate the effect of PGSM and systematically study robustness across problem dimensions and buffer sizes. These experiments provide insight into the optimizer’s convergence behavior and transfer to unseen problem dimensions. We then evaluate L-SR1 on a suite of benchmark optimization problems \citep{simulationlib} (Section~\ref{sec:profiles}), demonstrating consistent improvements over classical and learned baselines across diverse objective landscapes. Finally, we apply L-SR1 to a real-world 3D human mesh recovery task (Section~\ref{sec:hmr}), highlighting its effectiveness in high-dimensional, structured optimization settings; there we compare against LGD~\citep{song2020human} and an L-SR1 variant trained without the secant penalty, which we treat as representative learned-optimization baselines in this pipeline.
Further results and implementation details are given in Appendix~\ref{app:exps}.

\paragraph{Implementation Details.} 
Our method is implemented entirely in PyTorch~\citep{pytorch}\footnote{Project page and code: \url{https://gallif.github.io/lsr1/}.}.  
Following standard learned optimization practices~\citep{metz2022practical,metz2022velo,chen2022learning}, our training setup consists of inner optimization loops and outer meta-iterations. Using PyTorch's autograd framework, we compute gradients of the inner objective with respect to the optimization variables during each unrolled step, and gradients of the meta-loss (Eq.~(\ref{eq:loss})) with respect to the optimizer's parameters during each meta-iteration.
We use the AdamW~\citep{loshchilov2017decoupled} optimizer for meta-training, with a fixed learning rate of $10^{-4}$, momentum parameters $\beta_1 = 0.9$, $\beta_2 = 0.999$, and weight decay coefficient $\lambda = 0.01$. 
Unless otherwise stated, meta-training is run for 10K iterations and takes approximately 15 hours on a single NVIDIA GeForce RTX 3090 GPU.
For non-learned baselines, learning rates were selected by grid search on each task (details in Appendix~\ref{ap:implementations}).
Per-iteration runtime and memory are reported in Appendix~\ref{sec:compute}.

\subsection{Analytic Experiments}

\subsubsection{Quadratic Functions}
\label{sec:quadratic}
We begin our evaluation with randomly generated quadratic functions, which provide a controlled and analytically well-understood setting for studying optimizer behavior. This setting allows us to isolate the effect of individual design choices in L-SR1 in a controlled manner. In particular, we use quadratic objectives to address three questions:
(i) how PGSM affects optimization dynamics;
(ii) whether performance generalizes across problem dimensions not seen during training; and
(iii) how robust the optimizer is to variations in buffer size during training and inference.
By answering these questions in a simple setting, we aim to build intuition for the behavior of L-SR1 before evaluating it on more complex objectives.
We also compare two separately meta-trained variants, one with a scalar learning rate and one with an element-wise learning rate, on both quadratics and the HMR task; results are provided in Appendix~\ref{sec:lr_param} (Table~\ref{ta:lr_study}).

We consider quadratic functions of the form
\begin{equation}
f(\mathbf{x}) = \frac{1}{2} \mathbf{x}^\top \mathbf{H} \mathbf{x} + \mathbf{b}^\top \mathbf{x},
\end{equation}
where $\mathbf{H} \in \mathbb{R}^{N \times N}$ is a random positive semi-definite matrix and $\mathbf{b} \in \mathbb{R}^N$ is a random vector. For each experiment, we generate independent collections of such objectives together with corresponding initial points $\mathbf{x}_0$.

Unless otherwise stated, meta-training and validation are performed on quadratic functions with dimension $N=2$. A fixed validation set of 32 functions and initial points is used throughout training, while training batches consist of 128 randomly generated functions per meta-iteration. To evaluate generalization, we additionally construct a test set of 32 previously unseen functions with $N=10$. Further implementation details are provided in the Appendix~\ref{sec:app_quad}.

\begin{figure}[t]
\centering
\includegraphics[width=0.45\textwidth]{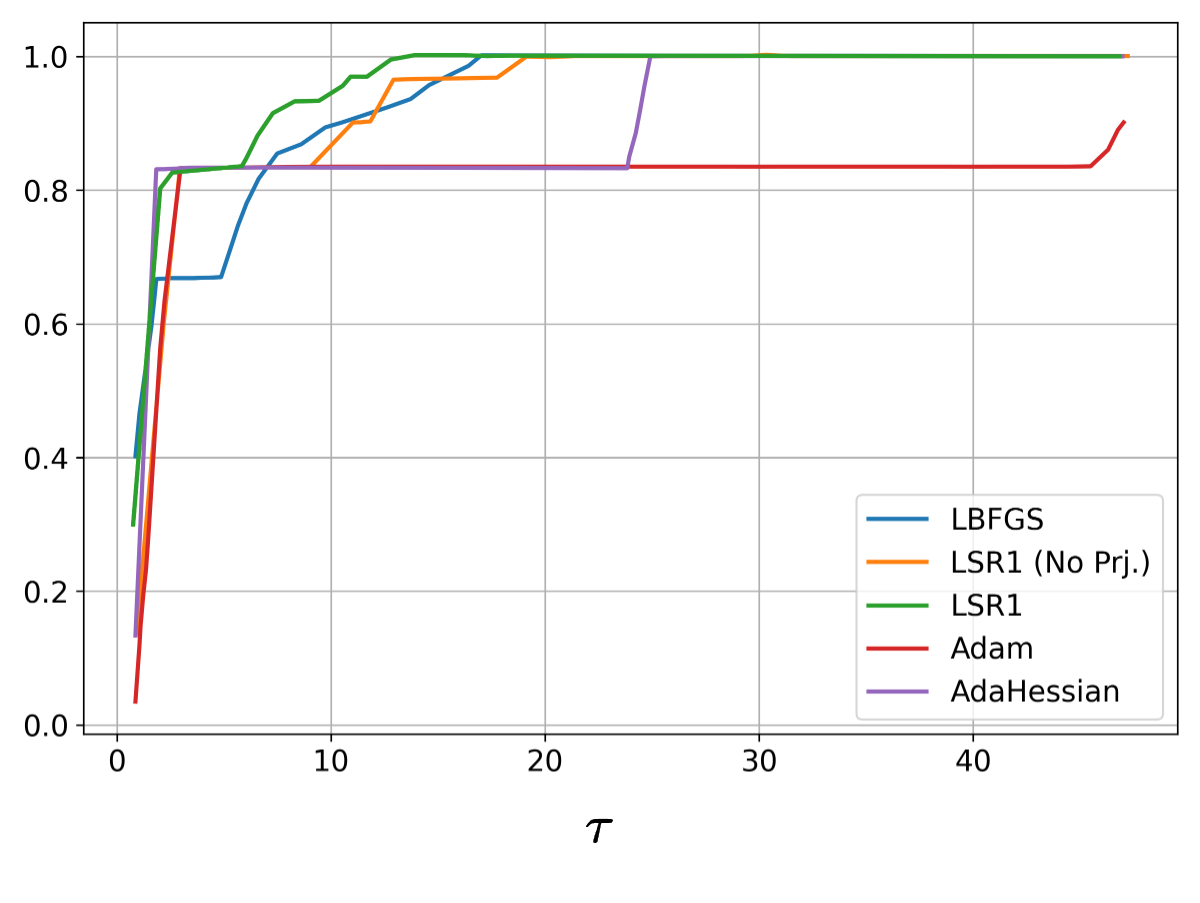}
\caption{\textbf{Performance profiles} comparing L-SR1 with PGSM and without (No Prj.) to baseline optimizers on a benchmark suite of objective functions. The benchmarks consist of quadratic, Rosenbrock, and Rastrigin functions across a range of problem dimensions. L-SR1 with PGSM achieves the highest overall performance profile, indicating strong effectiveness across tasks.}
\label{fig:profiles}
\end{figure}

\paragraph{Effect of PGSM.}
We first study the effect of PGSM in a controlled quadratic setting. 
Here all models were meta-trained using a fixed buffer size $L=8$ and evaluated using buffer sizes $L\in\{2,4,8\}$.
Figure~\ref{fig:vals} reports the average test loss over the first 20 optimization iterations. As expected, L-BFGS, which is data-independent and explicitly designed for quadratic problems, exhibits the fastest convergence across all tested settings. 
Importantly, L-SR1 with PGSM ($\lambda_{\text{sec}}>0$) consistently outperforms the variant without PGSM (No Prj., $\lambda_{\text{sec}}=0$), highlighting the benefit of the secant penalty during meta-training.
Despite being trained in low dimension, L-SR1 with PGSM remains stable and continues to make consistent progress when evaluated on higher-dimensional objectives.

To further examine the quality of the learned updates, Figure~\ref{fig:dirs} shows the average cosine similarity between the optimizer’s descent directions and the corresponding Newton directions on the test set. L-SR1 with PGSM exhibits steadily increasing alignment throughout optimization, indicating that it learns curvature-aware updates that generalize beyond the training dimension. 
In contrast, the variant without PGSM (No Prj.) exhibits weaker and less consistent alignment, highlighting the role of the secant penalty in shaping the learned updates.

\paragraph{Generalization Across Dimensions.}
Appendix~\ref{app:exps_stabilty} (Figure~\ref{fig:dim_stability_heatmap}) evaluates optimizers meta-trained only on $N{=}2$ quadratics on held-out test problems at $N{=}10$, with no re-training or dimension-specific retuning.
The heatmaps report mean-loss AUC over a fixed budget of 50 inner steps, swept over meta-training unroll depth and $\lambda_{\text{sec}}$.
Absolute AUC is higher at $N{=}10$ than at $N{=}2$ (e.g., L-SR1 with PGSM at unroll $16$ and $L{=}8$: $3.07 \pm 0.18$ vs.\ $7.33 \pm 0.26$ over four seeds). This is expected under a fixed iteration budget, as higher-dimensional problems require reducing more coordinates and the optimizer was never meta-trained on $N{=}10$ instances. Moreover, AUC integrates loss over the full optimization trajectory, so slower progress increases the score even when optimization remains stable, with no divergence observed in our runs.
More importantly, the ranking of hyperparameter settings is largely preserved across dimensions: configurations with intermediate unroll depths that perform best at $N{=}2$ remain among the best at $N{=}10$. This suggests that the method generalizes well across dimensions, and that models trained in lower-dimensional settings transfer meaningfully to higher-dimensional problems.

\paragraph{Buffer Sizes.}  
We evaluate the learned optimizer across training buffers $L_{\text{train}} \in \{2, 4, 8, 16\}$ and evaluation buffers $L_{\text{eval}} \in \{2, 4, 8, 16, 32, 64\}$ (Figure~\ref{fig:buffers}), and report results on the test set of previously unseen problems with dimension $N=10$. 
The area under the loss curve (AUC) is measured for a fixed iteration budget of 50 and averaged across random seeds.
Performance consistently improves with larger evaluation buffers, showing smooth, monotonic gains and reduced variance when additional curvature information is available at inference. For any fixed evaluation buffer, differences across training buffers are small and standard deviations remain modest, with no evidence of catastrophic mismatches.
Appendix~\ref{sec:compute} (Figs.~\ref{fig:comp_eval} and~\ref{fig:comp_train}) shows that meta-training incurs substantial memory cost due to unrolled graphs, which increases with buffer size and number of unrolled iterations. In contrast, inference is memory- and time-bounded with fixed per-iteration cost.

\subsubsection{Performance Profiles} \label{sec:profiles}

\begin{figure*}[t]
\centering
\begin{subfigure}[b]{0.48\textwidth}
    \centering
    \includegraphics[height=0.19\textheight]{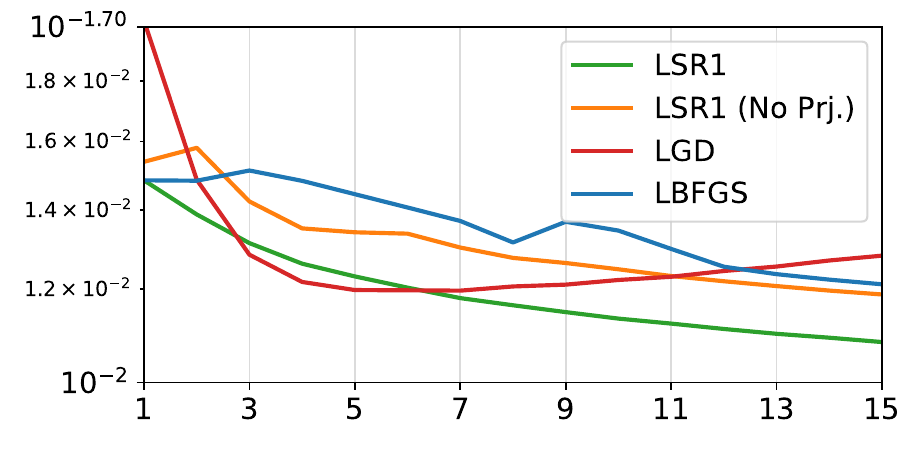}
    \caption{\textbf{2D joint reprojection error vs. update steps.}}
    \label{fig:hmr_repr}
\end{subfigure}
\hfill
\begin{subfigure}[b]{0.48\textwidth}
    \centering
    \includegraphics[height=0.19\textheight]{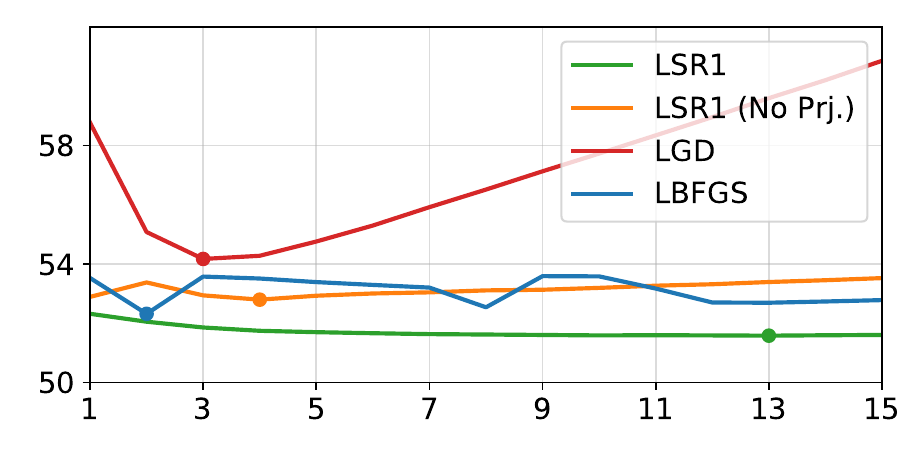}
    \caption{\textbf{PA-MPJPE vs. update steps.}}
    \label{fig:hmr_errors}
\end{subfigure}
\caption{\textbf{HMR error curves on 3DPW.} We compare 15 inner iterations of LGD~\citep{song2020human}, L-SR1 with PGSM (ours), L-SR1 without (No Prj.), and LBFGS~\citep{liu1989limited} initialized with our learned initialization. Fig.~\ref{fig:hmr_repr} shows 2D joint reprojection error; Fig.~\ref{fig:hmr_errors} shows PA-MPJPE. In Fig.~\ref{fig:hmr_errors}, markers indicate the best error achieved within these 15 inner iterations for each method. While LGD briefly achieves lower 2D error, L-SR1 with PGSM attains the strongest 3D accuracy, consistently outperforms the No Prj. variant, and continues to improve across iterations.
}

\label{fig:hmr_curves}
\end{figure*}

We compare our learned optimizer against several baselines using performance profiles, a standard evaluation method for optimization algorithms \citep{dolan2002benchmarking, sergeyev2015deterministic, beiranvand2017best, chen2022learning, gartner2023transformer}.
Protocol details (definitions, objective suite, and hyperparameters) are given in Appendix~\ref{ap:profiles}.
We meta-train on random quadratics and on Rosenbrock and Rastrigin functions with \(N{=}100\), and evaluate on a 30-problem suite spanning quadratics and non-convex objectives with dimensions 50--1000.
We compare L-SR1 (with and without PGSM) to L-BFGS~\citep{nocedal1980updating}, Adam~\citep{kingma2014adam}, and AdaHessian~\citep{yao2021adahessian}, reporting performance after a fixed iteration budget $K$; per-iteration costs differ across solvers (Appendix~\ref{sec:compute}).
Figure~\ref{fig:profiles} shows that L-SR1 with PGSM achieves the highest performance profile.

\subsection{Monocular Human Mesh Recovery (HMR)} \label{sec:hmr}


We integrate L-SR1 into the learned optimization–based HMR framework of \citep{song2020human}, which combines a trainable initialization with gradient-descent–inspired update modules. Self-supervision during training replaces the need for manually engineered loss terms \citep{bogo2016keep,pavlakos2019expressive}. Using L-SR1 demonstrates efficient convergence and improved accuracy compared to existing optimization-based HMR pipelines, while requiring no task-specific fine-tuning and using a smaller model. HMR provides a high-dimensional, non-convex real-world scenario to study optimizer behavior in practice (see Appendix \ref{app:exps_limited} for reduced-data experiments). In our evaluation, LGD and L-SR1 without PGSM represent the main learned-optimization baselines: LGD follows the original learned first-order-style inner loop of this framework, while the latter uses the same L-SR1 parameterization but omits the secant penalty at meta-training ($\lambda_{\text{sec}}=0$; labeled \emph{w/o sec.} or No Prj. in tables and figures).


\paragraph{Training}
Our model follows the training scheme of \citep{song2020human} and is trained exclusively on the AMASS dataset \citep{AMASS}, which contains approximately 20M human meshes with ground-truth SMPL shape and pose parameters \( (\bb{\beta}_\text{gt}, \bb{\theta}_\text{gt}) \). Using the SMPL body model \citep{loper2023smpl}\footnote{Although more expressive body models exist \citep{pavlakos2019expressive, STAR:2020}, SMPL is adopted for its widespread use and ease of integration.}, we synthetically generate 3D meshes and joint locations, which are projected to obtain 2D keypoints \( \bb{x}_\text{gt} \). These synthetic 2D and 3D annotations provide the self-supervised training signals.
At each inner iteration \( k \), the optimizer estimates parameters \( (\bb{\beta}_k, \bb{\theta}_k) \), from which predicted 3D and 2D joints \( (\bb{X}_k, \bb{x}_k) \) are obtained. The inner objective is defined as the weighted 2D reprojection loss
\begin{equation}
    f_{\text{rec}} \left( \bb{x}_k \right) = \left\| \bb{w} \odot \left(\bb{x}_k - \bb{x}_\text{gt}\right) \right\|_1,
\end{equation}
where \( \bb{w} \) are confidence weights randomly sampled from a Bernoulli distribution during training, following \citep{song2020human}.

Incorporating the self-supervision strategy of \citep{song2020human} together with the secant penalty, the total meta-loss is
\begin{equation}
    \Lcal_{\text{meta}} = \sum_{k=1}^K \left( \lambda_{\text{2D}} f\left( \bb{x}_k \right) + \lambda_{\text{self}} \|\Theta_k - \Theta_{\text{gt}}\|_1 \right) + \lambda_{\text{sec}}\mathcal{R}_{\text{sec}},
\end{equation}
where \( \Theta_k = \left\{ \bb{X}_k, \bb{\theta}_k, \bb{\beta}_k \right\} \) and \( \lambda_{\text{2D}} \), \( \lambda_{\text{self}} \), and \( \lambda_{\text{sec}} \) are scalar hyperparameters. Further implementation details are provided in Appendix~\ref{ap:hmr}.

\subsubsection{Evaluation on 3DPW}

\begin{table}[t]
\centering
\caption{\textbf{Evaluation on 3DPW.} PA-MPJPE of regression-based and optimization-based methods, with total inference time reported for learned optimization methods (timing details in Appendix~\ref{sec:compute}). Most approaches are trained on large multi-dataset combinations including in-the-wild data, while some methods are trained exclusively on AMASS \citep{AMASS}. Methods marked with $\dagger$ use 3DPW as part of their training data. L-SR1$_{\text{4-steps}}$ reports PA-MPJPE at the 4th inner iteration for comparison with LGD. L-SR1$_{\text{w/o sec.}}$ is a variant trained without the secant term.}

\label{ta:hmr_results}
\resizebox{0.48\textwidth}{!}{ 
\begin{tabular}{l | l | c | c | c}  
\toprule
 & \textbf{Method} & \textbf{PA-MPJPE} & \shortstack{\textbf{AMASS} \\ \textbf{only}} & \shortstack{\textbf{Time} \\ \textbf{(ms)}}\\
\midrule
\multirow{3}{*}{\rotatebox{90}{Regres.}} 
& TRACE$^\dagger$ \citep{sun2023trace} & 50.8 & \xmark & --\\
& WHAM (ViT)$^\dagger$ \citep{shin2024wham} & 35.9 & \xmark & --\\
& PromptHMR$^\dagger$ \citep{wang2025prompthmr} & 36.6 & \xmark & --\\
\midrule
\multirow{6}{*}{\rotatebox{90}{Optimization}} 
& SMPLify \citep{bogo2016keep} & 106.80 & No training & --\\
& SPIN \citep{kolotouros2019learning} & 59.20 & \xmark & --\\
& LGD \citep{song2020human} & 55.90  & \cmark & 664\\
\cmidrule{2-5} 
& L-SR1$_{\text{4-steps}}$ (Ours) & 51.74  & \cmark & \textbf{364}\\
& L-SR1$_{\text{w/o sec.}}$ (Ours) & 52.80  & \cmark & 364\\
& \textbf{L-SR1$_{\text{opt-horizon}}$ (Ours)} & \textbf{51.58} & \cmark & 1183\\
\bottomrule
\end{tabular}
}
\end{table}

We evaluate our method on the challenging 3DPW dataset \citep{vonMarcard2018}, which features complex, in-the-wild poses. We report PA-MPJPE on the test set, following the evaluation protocol of \citep{song2020human}, and using the 2D keypoints provided by OpenPose \citep{cao2019openpose}.
Table~\ref{ta:hmr_results} summarizes PA-MPJPE, training-data scope, and total inference time on 3DPW. L-SR1 outperforms LGD in accuracy with a smaller model and lower wall-clock cost at the 4-step horizon; the opt-horizon configuration trades additional inference time for the best reported PA-MPJPE. Fig.~\ref{fig:hmr_curves} shows the corresponding 2D reprojection inner loss and PA-MPJPE curves over optimization iterations.
A full per-iteration runtime and memory breakdown is given in Appendix~\ref{sec:compute}. Qualitative examples are shown in Fig.~\ref{fig:hmr_example}, with additional examples available in Appendix~\ref{ap:hmr_results}.

\begin{figure}[t]
\centering
\includegraphics[width=0.45\textwidth]{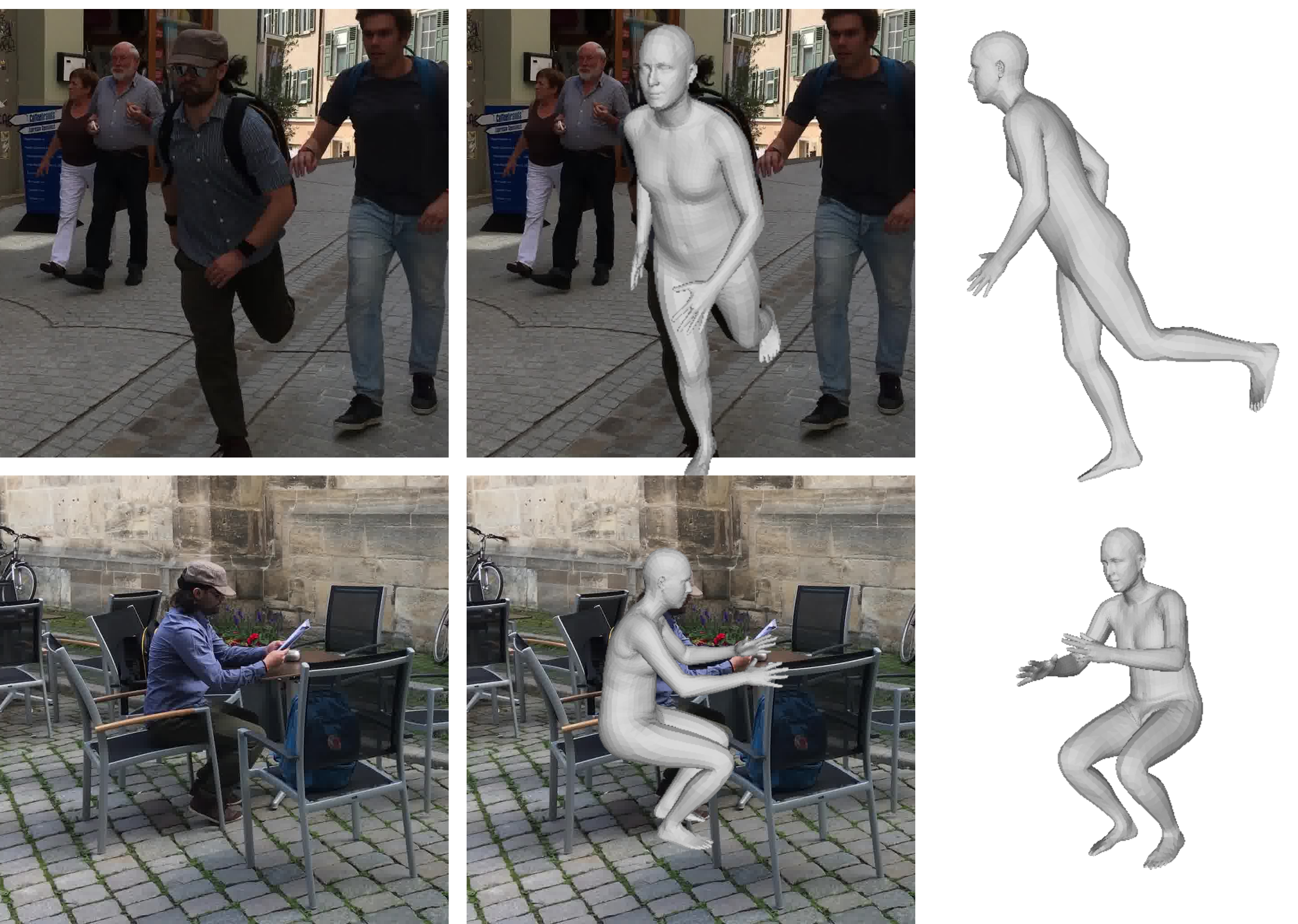}
\caption{\textbf{Qualitative examples from 3DPW.}}
\label{fig:hmr_example}
\end{figure}

\begin{figure}[t]
\centering
\includegraphics[width=\linewidth]{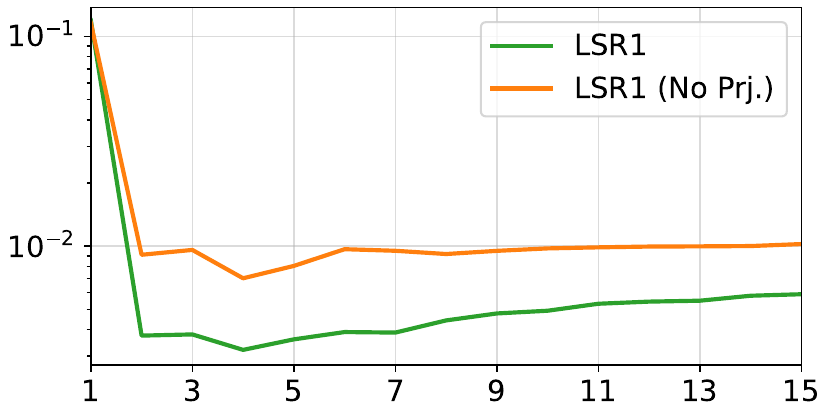}
\caption{\textbf{Pose-block secant penalty on HMR inner iterations (3DPW).} Average $r_k^{(\bb{\theta})}$ from \eqref{eq:hmr_secant_theta} over the 3DPW test set versus inner-update step $k$ for L-SR1 with PGSM ($\lambda_{\text{sec}}>0$) and without (No Prj.). The vertical axis uses a log scale; only the first 15 inner iterations are shown, matching Fig.~\ref{fig:hmr_curves}.}

\label{fig:hmr_secant_violation}
\end{figure}

\subsubsection{Secant penalty at inference}

Recall the secant penalty (Eq.~(\ref{eq:loss})) given in Section~\ref{sec:projection}. The corresponding difference terms are denoted as $\bb{p}_k = \bb{x}_k - \bb{x}_{k-1}$ and $\bb{q}_k = \bb{g}_k - \bb{g}_{k-1}$, where $\bb{x}_k$ and $\bb{g}_k$ are the inner-loop parameters and gradients, respectively.

In HMR, $\bb{x}_k$ includes SMPL parameters, in particular the pose $\bb{\theta}_k$ introduced above. Let $[\cdot]_{\bb{\theta}}$ denote the subvector corresponding to the pose block. We define the \emph{pose-block secant penalty} at iteration $k$ as
\begin{equation}
    r_k^{(\bb{\theta})} \;=\; \bigl\| \bigl[\bb{p}_k - \tilde{\bb{B}}_k \bb{q}_k\bigr]_{\bb{\theta}} \bigr\|_2^2\,.
    \label{eq:hmr_secant_theta}
\end{equation}

Fig.~\ref{fig:hmr_secant_violation} reports $r_k^{(\bb{\theta})}$ averaged over the 3DPW test set for the first 15 inner iterations. We focus on the $\bb{\theta}$ block as it is the highest-dimensional and most nonlinear component of the optimization, whereas the shape parameters $\bb{\beta}_k$ are comparatively low-dimensional.
We evaluate the same checkpoints as in Fig.~\ref{fig:hmr_curves}: L-SR1 with PGSM ($\lambda_{\text{sec}}>0$) and without (No Prj.). PGSM training yields substantially lower $r_k^{(\bb{\theta})}$, indicating improved secant consistency at test time.

\section{Conclusions}

We introduced L-SR1, a learned second-order optimizer built around PGSM, combining PSD-by-construction rank-one updates with secant-guided meta-training while incurring no additional projection cost at inference. On analytic benchmarks, L-SR1 with PGSM achieves the strongest overall performance profile, while on HMR it improves convergence and accuracy over competing learned optimization approaches using a smaller model and no task-specific fine-tuning. Our experiments focus on optimization-driven settings where each inner step is computationally expensive, such as HMR, where curvature-aware learned updates can reduce the number of costly iterations rather than serving as a drop-in optimizer for large-scale deep network training.
More broadly, L-SR1 can accelerate iterative gradient-based frameworks by reducing the number of optimization steps required for convergence. 

Limitations, including higher per-step cost than first-order methods and the scope of the empirical evaluation, are discussed in Appendix~\ref{sec:limits}.

\section*{Acknowledgements}
This study was funded partially by a scholarship from the Center for AI and Data Science at Tel Aviv University (TAD).

\section*{Impact Statement}

This paper presents work whose goal is to advance the field of Machine
Learning. There are many potential societal consequences of our work, none
which we feel must be specifically highlighted here.

\bibliography{egbib}

@book{boyd2004convex,
  title={Convex optimization},
  author={Boyd, Stephen and Boyd, Stephen P and Vandenberghe, Lieven},
  year={2004},
  publisher={Cambridge university press}
}

@article{nocedal1980updating,
  title={Updating quasi-Newton matrices with limited storage},
  author={Nocedal, Jorge},
  journal={Mathematics of computation},
  volume={35},
  number={151},
  pages={773--782},
  year={1980}
}

@article{shlezinger2023model,
  title={Model-based deep learning},
  author={Shlezinger, Nir and Whang, Jay and Eldar, Yonina C and Dimakis, Alexandros G},
  journal={Proceedings of the {IEEE}},
  volume={111},
  number={5},
  pages={698--720},
  year={2023},
  publisher={IEEE}
}

@incollection{pytorch,
title = {PyTorch: An Imperative Style, High-Performance Deep Learning Library},
author = {Paszke, Adam and Gross, Sam and Massa, Francisco and Lerer, Adam and Bradbury, James and Chanan, Gregory and Killeen, Trevor and Lin, Zeming and Gimelshein, Natalia and Antiga, Luca and Desmaison, Alban and Kopf, Andreas and Yang, Edward and DeVito, Zachary and Raison, Martin and Tejani, Alykhan and Chilamkurthy, Sasank and Steiner, Benoit and Fang, Lu and Bai, Junjie and Chintala, Soumith},
booktitle = {Advances in Neural Information Processing Systems 32},
editor = {H. Wallach and H. Larochelle and A. Beygelzimer and F. d\textquotesingle Alch\'{e}-Buc and E. Fox and R. Garnett},
pages = {8024--8035},
year = {2019},
publisher = {Curran Associates, Inc.},
url = {http://papers.neurips.cc/paper/9015-pytorch-an-imperative-style-high-performance-deep-learning-library.pdf}
}

@book{Bertsekas/99,
  author = {Bertsekas, D.P.},
  publisher = {Athena Scientific},
  title = {Nonlinear Programming},
  year = 1999
}

@article{conn1991convergence,
  title={Convergence of quasi-Newton matrices generated by the symmetric rank one update},
  author={Conn, Andrew R and Gould, Nicholas IM and Toint, Ph L},
  journal={Mathematical programming},
  volume={50},
  number={1},
  pages={177--195},
  year={1991},
  publisher={Springer}
}

@article{khalfan1993theoretical,
  title={A theoretical and experimental study of the symmetric rank-one update},
  author={Khalfan, H Fayez and Byrd, Richard H and Schnabel, Robert B},
  journal={SIAM Journal on Optimization},
  volume={3},
  number={1},
  pages={1--24},
  year={1993},
  publisher={SIAM}
}

@article{liu1989limited,
  title={On the limited memory {BFGS} method for large scale optimization},
  author={Liu, Dong C and Nocedal, Jorge},
  journal={Mathematical Programming},
  volume={45},
  number={1-3},
  pages={503--528},
  year={1989},
  publisher={Springer}
}

@inproceedings{nesterov1983method,
  title={A method of solving a convex programming problem with convergence rate O$\backslash$bigl(k\^{}2$\backslash$bigr)},
  author={Nesterov, Yurii Evgen'evich},
  booktitle={Doklady Akademii Nauk},
  volume={269},
  number={3},
  pages={543--547},
  year={1983},
  organization={Russian Academy of Sciences}
}

@inproceedings{metz2022practical,
  title={Practical tradeoffs between memory, compute, and performance in learned optimizers},
  author={Metz, Luke and Freeman, C Daniel and Harrison, James and Maheswaranathan, Niru and Sohl-Dickstein, Jascha},
  booktitle={Conference on Lifelong Learning Agents},
  pages={142--164},
  year={2022},
  organization={PMLR}
}

@inproceedings{metz2022velo,
  title={{VeLO}: Training Versatile Learned Optimizers by Scaling Up},
  author={Metz, Luke and Harrison, James and Freeman, C Daniel and Merchant, Amil and Beyer, Lucas and Bradbury, James and Agrawal, Naman and Poole, Ben and Mordatch, Igor and Roberts, Adam and Sohl-Dickstein, Jascha},
  booktitle={International Conference on Machine Learning},
  pages={17936--17957},
  year={2023},
  organization={PMLR}
}

@inproceedings{gartner2023transformer,
  title={Transformer-based learned optimization},
  author={G{\"a}rtner, Erik and Metz, Luke and Andriluka, Mykhaylo and Freeman, C Daniel and Sminchisescu, Cristian},
  booktitle={Proceedings of the IEEE/CVF Conference on Computer Vision and Pattern Recognition},
  pages={11970--11979},
  year={2023}
}

@article{chen2022learning,
  title={Learning to optimize: A primer and a benchmark},
  author={Chen, Tianlong and Chen, Xiaohan and Chen, Wuyang and Heaton, Howard and Liu, Jialin and Wang, Zhangyang and Yin, Wotao},
  journal={Journal of Machine Learning Research},
  volume={23},
  number={189},
  pages={1--59},
  year={2022}
}

@MISC{simulationlib, 
 author = {Surjanovic, S. and Bingham, D.}, 
 title = {Virtual Library of Simulation Experiments: Test Functions and Datasets}, 
 howpublished = {Retrieved May 5, 2024, from \url{http://www.sfu.ca/~ssurjano}} 
}

@inproceedings{sutskever2013importance,
  title={On the importance of initialization and momentum in deep learning},
  author={Sutskever, Ilya and Martens, James and Dahl, George and Hinton, Geoffrey},
  booktitle={International conference on machine learning},
  pages={1139--1147},
  year={2013},
  organization={PMLR}
}

@inproceedings{martens2015optimizing,
  title={Optimizing neural networks with kronecker-factored approximate curvature},
  author={Martens, James and Grosse, Roger},
  booktitle={International conference on machine learning},
  pages={2408--2417},
  year={2015},
  organization={PMLR}
}

@inproceedings{gupta2018shampoo,
  title={Shampoo: Preconditioned stochastic tensor optimization},
  author={Gupta, Vineet and Koren, Tomer and Singer, Yoram},
  booktitle={International Conference on Machine Learning},
  pages={1842--1850},
  year={2018},
  organization={PMLR}
}

@article{andrychowicz2016learning,
  title={Learning to learn by gradient descent by gradient descent},
  author={Andrychowicz, Marcin and Denil, Misha and Gomez, Sergio and Hoffman, Matthew W and Pfau, David and Schaul, Tom and Shillingford, Brendan and De Freitas, Nando},
  journal={Advances in neural information processing systems},
  volume={29},
  year={2016}
}

@article{li2016learning,
  title={Learning to optimize},
  author={Li, Ke and Malik, Jitendra},
  journal={arXiv preprint arXiv:1606.01885},
  year={2016}
}

@article{metz2020tasks,
  title={Tasks, stability, architecture, and compute: Training more effective learned optimizers, and using them to train themselves},
  author={Metz, Luke and Maheswaranathan, Niru and Freeman, C Daniel and Poole, Ben and Sohl-Dickstein, Jascha},
  journal={arXiv preprint arXiv:2009.11243},
  year={2020}
}

@inproceedings{wichrowska2017learned,
  title={Learned optimizers that scale and generalize},
  author={Wichrowska, Olga and Maheswaranathan, Niru and Hoffman, Matthew W and Colmenarejo, Sergio Gomez and Denil, Misha and Freitas, Nando and Sohl-Dickstein, Jascha},
  booktitle={International conference on machine learning},
  pages={3751--3760},
  year={2017},
  organization={PMLR}
}

@article{beiranvand2017best,
  title={Best practices for comparing optimization algorithms},
  author={Beiranvand, Vahid and Hare, Warren and Lucet, Yves},
  journal={Optimization and Engineering},
  volume={18},
  pages={815--848},
  year={2017},
  publisher={Springer}
}

@article{kingma2014adam,
  title={Adam: A method for stochastic optimization},
  author={Kingma, Diederik P and Ba, Jimmy},
  journal={arXiv preprint arXiv:1412.6980},
  year={2014}
}

@article{shlezinger2020viterbinet,
  title={ViterbiNet: A deep learning based Viterbi algorithm for symbol detection},
  author={Shlezinger, Nir and Farsad, Nariman and Eldar, Yonina C and Goldsmith, Andrea J},
  journal={IEEE Transactions on Wireless Communications},
  volume={19},
  number={5},
  pages={3319--3331},
  year={2020},
  publisher={IEEE}
}

@article{revach2022kalmannet,
  title={KalmanNet: Neural network aided Kalman filtering for partially known dynamics},
  author={Revach, Guy and Shlezinger, Nir and Ni, Xiaoyong and Escoriza, Adria Lopez and Van Sloun, Ruud JG and Eldar, Yonina C},
  journal={IEEE Transactions on Signal Processing},
  volume={70},
  pages={1532--1547},
  year={2022},
  publisher={IEEE}
}

@inproceedings{li2020learning,
  title={Learning to Combine Quasi-Newton Methods},
  author={Li, Maojia and Liu, Jialin and Yin, Wotao},
  booktitle={OPT2020: 12th Annual Workshop on Optimization for Machine Learning, NeurIPS},
  year={2020},
  url={https://opt-ml.org/papers/2020/paper_87.pdf}
}

@misc{liao2023learningoptimizequasinewtonmethods,
  title={Learning to Optimize Quasi-Newton Methods},
  author={Liao, Isaac and Dangovski, Rumen R. and Foerster, Jakob N. and Solja{\v{c}}i{\'c}, Marin},
  year={2023},
  eprint={2210.06171},
  archivePrefix={arXiv},
  primaryClass={cs.LG},
  url={https://arxiv.org/abs/2210.06171}
}

@inproceedings{yao2021adahessian,
  title={Adahessian: An adaptive second order optimizer for machine learning},
  author={Yao, Zhewei and Gholami, Amir and Shen, Sheng and Mustafa, Mustafa and Keutzer, Kurt and Mahoney, Michael},
  booktitle={proceedings of the AAAI conference on artificial intelligence},
  volume={35},
  number={12},
  pages={10665--10673},
  year={2021}
}

@article{loshchilov2017decoupled,
  title={Decoupled weight decay regularization},
  author={Loshchilov, Ilya and Hutter, Frank},
  journal={arXiv preprint arXiv:1711.05101},
  year={2017}
}

@article{dolan2002benchmarking,
  title={Benchmarking optimization software with performance profiles},
  author={Dolan, Elizabeth D and Mor{\'e}, Jorge J},
  journal={Mathematical programming},
  volume={91},
  pages={201--213},
  year={2002},
  publisher={Springer}
}

@article{sergeyev2015deterministic,
  title={A deterministic global optimization using smooth diagonal auxiliary functions},
  author={Sergeyev, Yaroslav D and Kvasov, Dmitri E},
  journal={Communications in Nonlinear Science and Numerical Simulation},
  volume={21},
  number={1-3},
  pages={99--111},
  year={2015},
  publisher={Elsevier}
}

@inproceedings{song2020human,
  title={Human body model fitting by learned gradient descent},
  author={Song, Jie and Chen, Xu and Hilliges, Otmar},
  booktitle={European Conference on Computer Vision},
  pages={744--760},
  year={2020},
  organization={Springer}
}

@incollection{loper2023smpl,
  title={SMPL: A skinned multi-person linear model},
  author={Loper, Matthew and Mahmood, Naureen and Romero, Javier and Pons-Moll, Gerard and Black, Michael J},
  booktitle={Seminal Graphics Papers: Pushing the Boundaries, Volume 2},
  pages={851--866},
  year={2023}
}

@conference{AMASS,
  title           = {{AMASS}: Archive of Motion Capture as Surface Shapes},
  author          = {Mahmood, Naureen and Ghorbani, Nima and Troje, Nikolaus F. and Pons-Moll, Gerard and Black, Michael J.},
  booktitle       = {International Conference on Computer Vision},
  pages           = {5442--5451},
  month           = oct,
  year            = {2019},
  month_numeric   = {10}
}

@misc{AMASS_ACCAD,
  title           = {{ACCAD MoCap Dataset}},
  author          = {{Advanced Computing Center for the Arts and Design}},
  url             = {https://accad.osu.edu/research/motion-lab/mocap-system-and-data}
}

@inproceedings{AMASS_BMLhandball,
  author          = {Helm, Fabian and Troje, Nikolaus and Reiser, Mathias and Munzert, J{\"o}rn},
  year            = {2015},
  month           = {01},
  pages           = {},
  title           = {Bewegungsanalyse get{\"a}uschter und nicht-get{\"a}uschter 7m-W{\"u}rfe im Handball},
  booktitle       = {47. Jahrestagung der Arbeitsgemeinschaft f{\"u}r Sportpsychologie, Freiburg}
}

@article{AMASS_BMLmovi,
  title           = {{MoVi}: A Large Multipurpose Motion and Video Dataset},
  author          = {Saeed Ghorbani and Kimia Mahdaviani and Anne Thaler and Konrad Kording and Douglas James Cook and Gunnar Blohm and Nikolaus F. Troje},
  year            = {2020},
  journal         = {arXiv preprint arXiv: 2003.01888}
}

@article{AMASS_BMLrub,
  title           = {Decomposing Biological Motion: {A} Framework for Analysis and Synthesis of Human Gait Patterns},
  author          = {Troje, Nikolaus F.},
  year            = 2002,
  month           = sep,
  journal         = {Journal of Vision},
  volume          = 2,
  number          = 5,
  pages           = {2--2},
  doi             = {10.1167/2.5.2},
  month_numeric   = 9
}

@misc{AMASS_CMU,
  title           = {{CMU MoCap Dataset}},
  author          = {{Carnegie Mellon University}},
  url             = {http://mocap.cs.cmu.edu}
}

@article{AMASS_DanceDB,
  author          = {Aristidou, Andreas and Shamir, Ariel and Chrysanthou, Yiorgos},
  title           = {Digital Dance Ethnography: {O}rganizing Large Dance Collections},
  journal         = {J. Comput. Cult. Herit.},
  issue_date      = {January 2020},
  volume          = {12},
  number          = {4},
  month           = nov,
  year            = {2019},
  issn            = {1556-4673},
  articleno       = {29},
  numpages        = {27},
  url             = {https://doi.org/10.1145/3344383},
  doi             = {10.1145/3344383},
  acmid           = {},
  publisher       = {Association for Computing Machinery},
  address         = {New York, NY, USA},
}

@inproceedings{AMASS_DFaust,
  title           = {Dynamic {FAUST}: {R}egistering Human Bodies in Motion},
  author          = {Bogo, Federica and Romero, Javier and Pons-Moll, Gerard and Black, Michael J.},
  booktitle       = {IEEE Conf. on Computer Vision and Pattern Recognition (CVPR)},
  month           = jul,
  year            = {2017},
  month_numeric   = {7}
}

@misc{AMASS_EyesJapanDataset,
  title           = {{Eyes Japan MoCap Dataset}},
  author          = {Eyes JAPAN Co. Ltd.},
  url             = {http://mocapdata.com}
}

@inproceedings{AMASS_GRAB,
  title           = {{GRAB}: A Dataset of Whole-Body Human Grasping of Objects},
  author          = {Taheri, Omid and Ghorbani, Nima and Black, Michael J. and Tzionas, Dimitrios},
  booktitle       = {European Conference on Computer Vision (ECCV)},
  year            = {2020},
  url             = {https://grab.is.tue.mpg.de}
}

@inproceedings{AMASS_GRAB-2,
  title           = {{ContactDB}: Analyzing and Predicting Grasp Contact via Thermal Imaging},
  author          = {Brahmbhatt, Samarth and Ham, Cusuh and Kemp, Charles C. and Hays, James},
  booktitle       = {The IEEE Conference on Computer Vision and Pattern Recognition (CVPR)},
  year            = {2019},
  url             = {https://contactdb.cc.gatech.edu}
}

@techreport{AMASS_HDM05,
  author          = {M. M\"{u}ller and T. R\"{o}der and M. Clausen and B. Eberhardt and B. Kr\"{u}ger and A. Weber},
  title           = {Documentation Mocap Database HDM05},
  number          = {CG-2007-2},
  year            = {2007},
  month           = {June},
  institution     = {Universit\"{a}t Bonn},
  issn            = {1610-8892}
}

@article{AMASS_HUMAN4D,
  title           = {HUMAN4D: A Human-Centric Multimodal Dataset for Motions and Immersive Media},
  author          = {Chatzitofis, Anargyros and Saroglou, Leonidas and Boutis, Prodromos and Drakoulis, Petros and Zioulis, Nikolaos and Subramanyam, Shishir and Kevelham, Bart and Charbonnier, Caecilia and Cesar, Pablo and Zarpalas, Dimitrios and others},
  journal         = {IEEE Access},
  volume          = {8},
  pages           = {176241--176262},
  year            = {2020},
  publisher       = {IEEE}
}

@article{AMASS_HumanEva,
  title           = {{HumanEva}: Synchronized video and motion capture dataset and baseline algorithm for evaluation of articulated human motion},
  author          = {Sigal, L. and Balan, A. and Black, M. J.},
  journal         = {International Journal of Computer Vision},
  volume          = {87},
  number          = {1},
  pages           = {4--27},
  publisher       = {Springer Netherlands},
  month           = mar,
  year            = {2010},
  doi             = {},
  month_numeric   = {3}
}

@inproceedings{AMASS_KIT-CNRS-EKUT-WEIZMANN,
  author          = {Christian Mandery and \"Omer Terlemez and Martin Do and Nikolaus Vahrenkamp and Tamim Asfour},
  title           = {The {KIT} Whole-Body Human Motion Database},
  booktitle       = {International Conference on Advanced Robotics (ICAR)},
  pages           = {329--336},
  year            = {2015},
}

@article{AMASS_KIT-CNRS-EKUT-WEIZMANN-2,
  author          = {Christian Mandery and \"Omer Terlemez and Martin Do and Nikolaus Vahrenkamp and Tamim Asfour},
  title           = {Unifying Representations and Large-Scale Whole-Body Motion Databases for Studying Human Motion},
  pages           = {796--809},
  volume          = {32},
  number          = {4},
  journal         = {IEEE Transactions on Robotics},
  year            = {2016},
}

@inproceedings{AMASS_KIT-CNRS-EKUT-WEIZMANN-3,
  author          = {Franziska Krebs and Andre Meixner and Isabel Patzer and Tamim Asfour},
  title           = {The {KIT} Bimanual Manipulation Dataset},
  booktitle       = {IEEE/RAS International Conference on Humanoid Robots (Humanoids)},
  pages           = {499--506},
  year            = {2021},
}

@inproceedings{AMASS_MOYO,
  title           = {{3D} Human Pose Estimation via Intuitive Physics},
  author          = {Tripathi, Shashank and M{\"u}ller, Lea and Huang, Chun-Hao P. and Taheri Omid and Black, Michael J. and Tzionas, Dimitrios},
  booktitle       = {Proceedings of the IEEE/CVF Conference on Computer Vision and Pattern Recognition (CVPR)},
  month           = {June},
  year            = {2023}
}

@article{AMASS_MoSh,
  title           = {{MoSh}: Motion and Shape Capture from Sparse Markers},
  author          = {Loper, Matthew M. and Mahmood, Naureen and Black, Michael J.},
  address         = {New York, NY, USA},
  publisher       = {ACM},
  month           = nov,
  number          = {6},
  volume          = {33},
  pages           = {220:1--220:13},
  journal         = {ACM Transactions on Graphics, (Proc. SIGGRAPH Asia)},
  url             = {http://doi.acm.org/10.1145/2661229.2661273},
  year            = {2014},
  doi             = {10.1145/2661229.2661273}
}

@inproceedings{AMASS_PosePrior,
  title           = {Pose-Conditioned Joint Angle Limits for {3D} Human Pose Reconstruction},
  author          = {Akhter, Ijaz and Black, Michael J.},
  booktitle       = { IEEE Conf. on Computer Vision and Pattern Recognition (CVPR) 2015},
  month           = jun,
  year            = {2015}
}

@misc{AMASS_SFU,
  title           = {{SFU Motion Capture Database}},
  author          = {Simon Fraser University and National University of Singapore},
  url             = {http://mocap.cs.sfu.ca/}
}

@inproceedings{AMASS_SOMA,
  title           = {{SOMA}: Solving Optical Marker-Based MoCap Automatically},
  author          = {Ghorbani, Nima and Black, Michael J.},
  booktitle       = {Proc. International Conference on Computer Vision (ICCV)},
  pages           = {11117--11126},
  month           = oct,
  year            = {2021},
  doi             = {},
  month_numeric   = {10}
}

@inproceedings{AMASS_TCDHands,
  author          = {Ludovic Hoyet and Kenneth Ryall and Rachel McDonnell and Carol O'Sullivan},
  title           = {Sleight of Hand: Perception of Finger Motion from Reduced Marker Sets},
  booktitle       = {Proceedings of the ACM SIGGRAPH Symposium on Interactive 3D Graphics and Games},
  year            = {2012},
  pages           = {79--86},
  doi             = {10.1145/2159616.2159629}
}

@inproceedings{AMASS_TotalCapture,
  author          = {Trumble, Matt and Gilbert, Andrew and Malleson, Charles and  Hilton, Adrian and Collomosse, John},
  title           = {{Total Capture}: 3D Human Pose Estimation Fusing Video and Inertial Sensors},
  booktitle       = {2017 British Machine Vision Conference (BMVC)},
  year            = {2017}
}

@inproceedings{AMASS_WheelPoser,
  title={WheelPoser: Sparse-IMU Based Body Pose Estimation for Wheelchair Users},
  author={Li, Yunzhi and Mollyn, Vimal and Yuan, Kuang and Carrington, Patrick},
  booktitle={Proceedings of the 26th International ACM SIGACCESS Conference on Computers and Accessibility},
  pages={1--17},
  year={2024}
}

@inproceedings{pavlakos2019expressive,
  title={Expressive body capture: 3d hands, face, and body from a single image},
  author={Pavlakos, Georgios and Choutas, Vasileios and Ghorbani, Nima and Bolkart, Timo and Osman, Ahmed AA and Tzionas, Dimitrios and Black, Michael J},
  booktitle={Proceedings of the IEEE/CVF conference on computer vision and pattern recognition},
  pages={10975--10985},
  year={2019}
}

@inproceedings{STAR:2020,
      author = {Osman, Ahmed A A and Bolkart, Timo and Black, Michael J.},
      title = {{STAR}: A Sparse Trained Articulated Human Body Regressor},
      booktitle = {European Conference on Computer Vision (ECCV)},
      pages = {598--613},
      year = {2020},
      url = {https://star.is.tue.mpg.de}
}

@inproceedings{bogo2016keep,
  title={Keep it SMPL: Automatic estimation of 3D human pose and shape from a single image},
  author={Bogo, Federica and Kanazawa, Angjoo and Lassner, Christoph and Gehler, Peter and Romero, Javier and Black, Michael J},
  booktitle={Computer Vision--ECCV 2016: 14th European Conference, Amsterdam, The Netherlands, October 11-14, 2016, Proceedings, Part V 14},
  pages={561--578},
  year={2016},
  organization={Springer}
}

@inproceedings{kolotouros2019learning,
  title={Learning to reconstruct 3D human pose and shape via model-fitting in the loop},
  author={Kolotouros, Nikos and Pavlakos, Georgios and Black, Michael J and Daniilidis, Kostas},
  booktitle={Proceedings of the IEEE/CVF international conference on computer vision},
  pages={2252--2261},
  year={2019}
}

@inproceedings{shin2024wham,
  title={Wham: Reconstructing world-grounded humans with accurate 3d motion},
  author={Shin, Soyong and Kim, Juyong and Halilaj, Eni and Black, Michael J},
  booktitle={Proceedings of the IEEE/CVF Conference on Computer Vision and Pattern Recognition},
  pages={2070--2080},
  year={2024}
}

@inproceedings{sun2023trace,
  title={Trace: 5d temporal regression of avatars with dynamic cameras in 3d environments},
  author={Sun, Yu and Bao, Qian and Liu, Wu and Mei, Tao and Black, Michael J},
  booktitle={Proceedings of the IEEE/CVF Conference on Computer Vision and Pattern Recognition},
  pages={8856--8866},
  year={2023}
}

@article{wang2025prompthmr,
  title={PromptHMR: Promptable Human Mesh Recovery},
  author={Wang, Yufu and Sun, Yu and Patel, Priyanka and Daniilidis, Kostas and Black, Michael J and Kocabas, Muhammed},
  journal={arXiv preprint arXiv:2504.06397},
  year={2025}
}

@inproceedings{ayad2024qn,
  title={QN-Mixer: A Quasi-Newton MLP-Mixer Model for Sparse-View CT Reconstruction},
  author={Ayad, Ishak and Larue, Nicolas and Nguyen, Ma{\"\i} K},
  booktitle={Proceedings of the IEEE/CVF Conference on Computer Vision and Pattern Recognition},
  pages={25317--25326},
  year={2024}
}

@article{jain2023mnemosyne,
  title={Mnemosyne: Learning to train transformers with transformers},
  author={Jain, Deepali and Choromanski, Krzysztof M and Dubey, Kumar Avinava and Singh, Sumeet and Sindhwani, Vikas and Zhang, Tingnan and Tan, Jie},
  journal={Advances in Neural Information Processing Systems},
  volume={36},
  pages={77331--77358},
  year={2023}
}

@inproceedings{shetty2023pliks,
  title={Pliks: A pseudo-linear inverse kinematic solver for 3d human body estimation},
  author={Shetty, Karthik and Birkhold, Annette and Jaganathan, Srikrishna and Strobel, Norbert and Kowarschik, Markus and Maier, Andreas and Egger, Bernhard},
  booktitle={Proceedings of the IEEE/CVF conference on computer vision and pattern recognition},
  pages={574--584},
  year={2023}
}

@inproceedings{vonMarcard2018,
title = {Recovering Accurate 3D Human Pose in The Wild Using IMUs and a Moving Camera},
author = {von Marcard, Timo and Henschel, Roberto and Black, Michael and Rosenhahn, Bodo and Pons-Moll, Gerard},
booktitle = {European Conference on Computer Vision (ECCV)},
year = {2018},
month = {sep}
}

@article{cao2019openpose,
  title={Openpose: Realtime multi-person 2d pose estimation using part affinity fields},
  author={Cao, Zhe and Hidalgo, Gines and Simon, Tomas and Wei, Shih-En and Sheikh, Yaser},
  journal={IEEE transactions on pattern analysis and machine intelligence},
  volume={43},
  number={1},
  pages={172--186},
  year={2019},
  publisher={IEEE}
}
\bibliographystyle{icml2026}

\newpage
\appendix
\onecolumn
\section{Computational Analysis} \label{sec:compute}
\begin{figure}[h!]
\centering
\includegraphics[width=\textwidth]{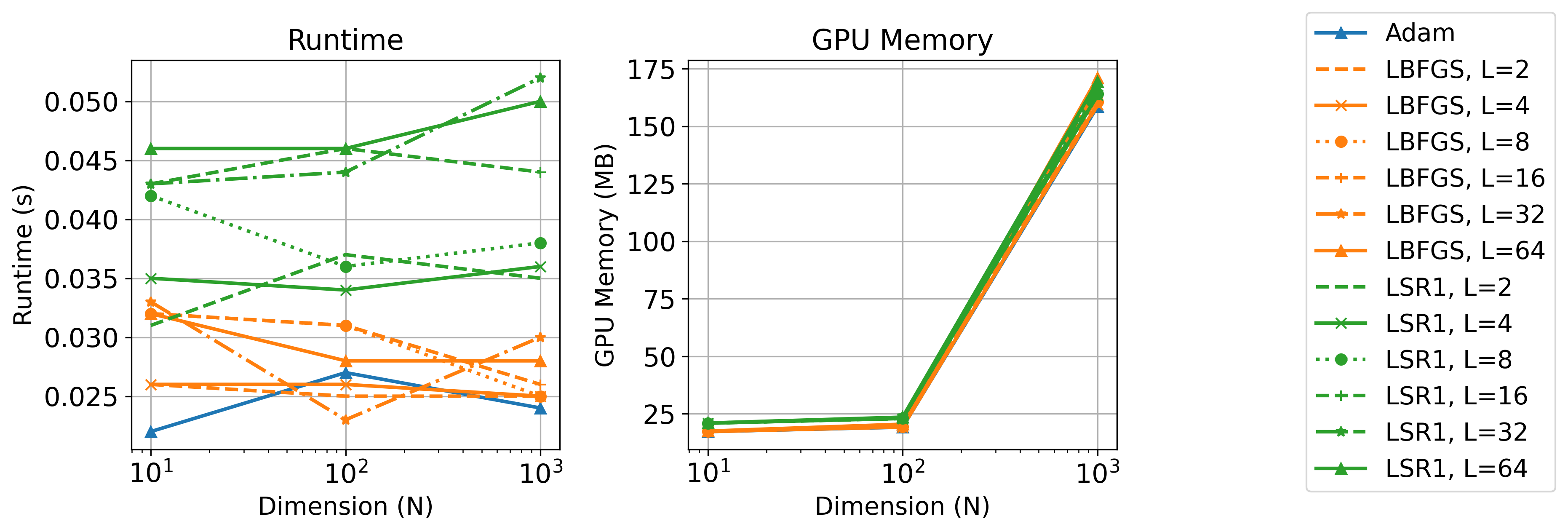}
\caption{
\textbf{Computational effort during inference.} Measurements were collected on a single NVIDIA RTX 3090 GPU and correspond 
to the mean runtime per inner inference iteration and peak memory with a batch size of 32.
}
\label{fig:comp_eval}
\end{figure}
\begin{figure}[h!]
\centering
\includegraphics[width=\textwidth]{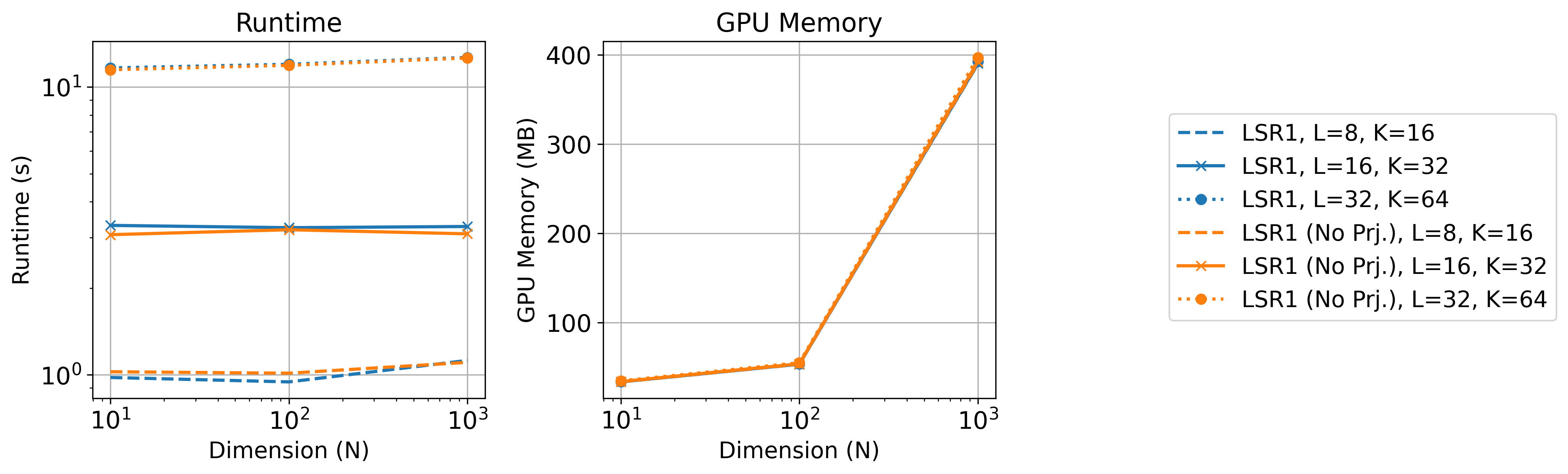}
\caption{
\textbf{Computational effort during meta-training.} Measurements were collected on a single NVIDIA RTX 3090 GPU and correspond 
to the mean runtime per inner inference iteration and peak memory
usage with a batch size of 4.
}
\label{fig:comp_train}
\end{figure}
\begin{table}[h!]
  \centering
  \caption{\textbf{HMR runtime and memory comparison.} Measurements were collected on a single NVIDIA RTX 3090 GPU and correspond to the mean runtime per inner inference iteration and peak memory usage, using a batch size of 256 and buffer size $L=4$ for L-SR1.}
  \label{tab:runtime_memory}
  \begin{tabular}{lccc}
    \toprule
    \textbf{Method} & \textbf{Runtime (ms)} & \textbf{Memory (GiB)} & \textbf{\# Params}\\
    \midrule
    LGD \citep{song2020human}  & 166   & 17.81 & 17.4 M\\
    L-SR1 & 91    & 14.60  & 10.4 M\\
    \bottomrule
  \end{tabular}
\end{table}

\newpage
\section{Further Studies and Analyses} \label{app:exps}

\subsection{Generalization Across Dimensions} \label{app:exps_stabilty}
\begin{figure*}[h]
    \centering
    \begin{subfigure}[t]{0.48\textwidth}
        \centering
        \includegraphics[width=\textwidth]{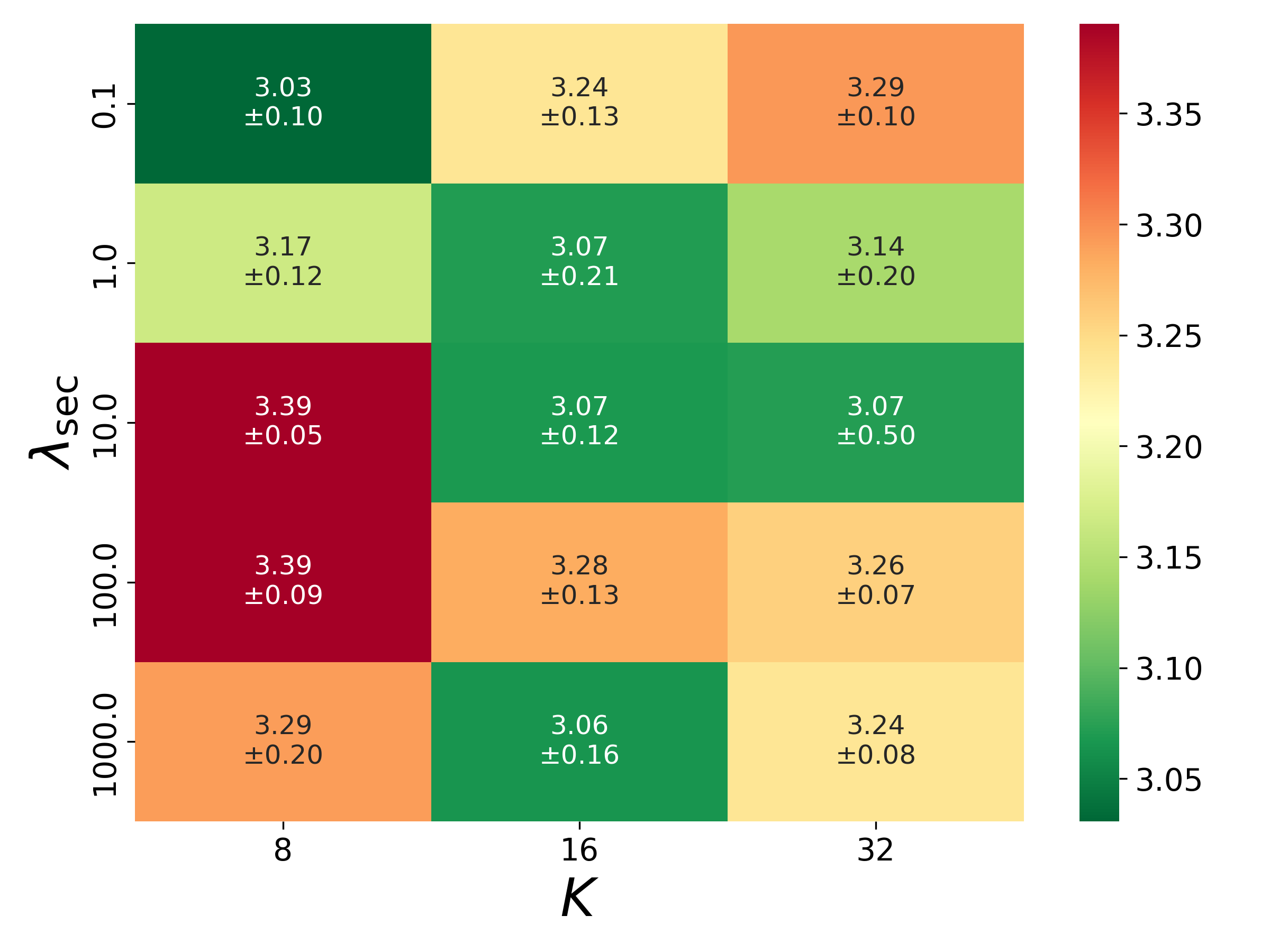}
        \caption{$N=2$ (training dimension)}
        \label{fig:dim_stability_dim2}
    \end{subfigure}
    \hfill
    \begin{subfigure}[t]{0.48\textwidth}
        \centering
        \includegraphics[width=\textwidth]{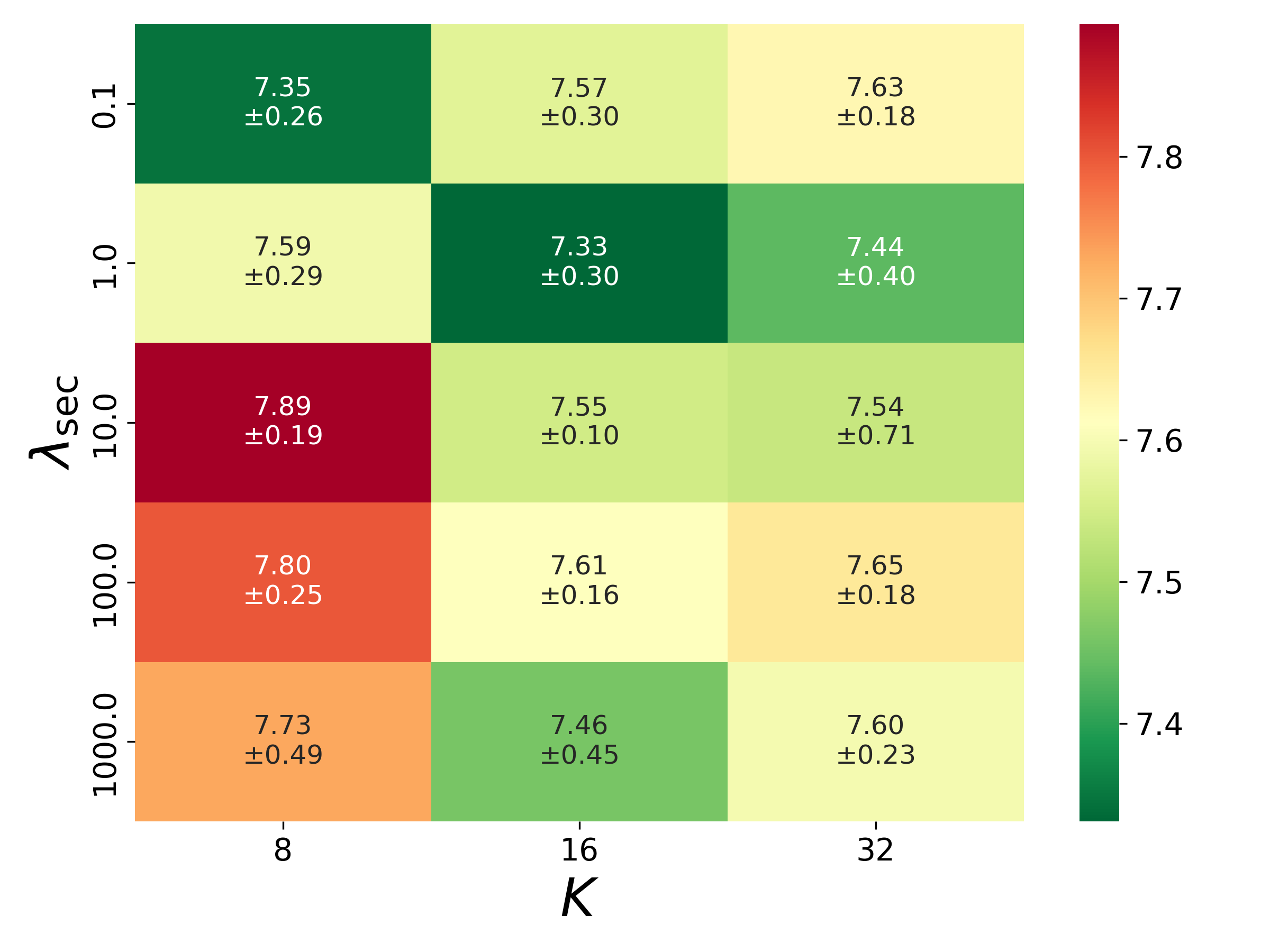}
        \caption{$N=10$ (unseen test dimension)}
        \label{fig:dim_stability_dim10}
    \end{subfigure}

    \caption{
    \textbf{Generalization across dimensions.}
    Heatmaps report the area under the loss curve (AUC) for the learned optimizer across combinations of unrolling depth and PGSM secant-penalty weight $\lambda_{\text{sec}}$.
    The optimizer is meta-trained exclusively on quadratic objectives with $N=2$ and evaluated without re-training on unseen test problems with $N=2$ (left) and $N=10$ (right).
    Mean-loss AUC is higher at $N=10$ under a fixed 50-step budget, but hyperparameter rankings (e.g., intermediate unroll depths) are largely preserved across dimensions.
    }
    \label{fig:dim_stability_heatmap}
\end{figure*}

This appendix studies whether L-SR1 meta-trained on low-dimensional quadratics can be applied at a higher test dimension without re-training.

\paragraph{Experimental setup.}
All models are meta-trained on randomly generated quadratic objectives with $N=2$ (Section~\ref{sec:quadratic}) and evaluated on held-out quadratics with $N=10$.
No re-training or dimension-specific hyperparameter tuning is performed at test time.
We sweep meta-training unroll depth and the PGSM secant-penalty weight $\lambda_{\text{sec}}$, with four random seeds per configuration.
Performance is summarized by the area under the mean loss curve (AUC) over 50 inner optimization steps.

\paragraph{Results.}
Figure~\ref{fig:dim_stability_heatmap} compares the resulting heatmaps at $N=2$ (training dimension) and $N=10$ (unseen test dimension).
Absolute AUC increases at $N=10$ under the fixed step budget: with more variables per problem and meta-training confined to $N=2$, the optimizer typically reaches higher time-averaged loss within 50 steps, even though trajectories remain well-behaved and we observe no systematic divergence.
The qualitative picture across hyperparameters is nevertheless similar: intermediate unroll depths outperform very short or very long unrolls at both dimensions, so configurations that work well after meta-training at $N=2$ remain competitive at $N=10$ without dimension-specific retuning.

\subsection{Learning-Rate Scalarization} \label{sec:lr_param}
L-SR1 is defined with an element-wise rate vector $\boldsymbol{\alpha}_k$ from $\mathcal{G}$ (Section~\ref{sec:componenets}), and all primary experiments use that design. Here we isolate how much coordinate-wise flexibility matters by meta-training \textbf{distinct} checkpoints under an alternative inner loop: the same $\mathcal{G}$ is used, but after
\[
\boldsymbol{\alpha}_k = \gamma_1 \cdot \exp\left( \gamma_2 \cdot \tilde{\boldsymbol{\alpha}}_k \right),
\]
we replace $\boldsymbol{\alpha}_k$ with its coordinate average, broadcast to all parameters,
\begin{equation}
\label{eq:lr_scalarize}
\boldsymbol{\alpha}_k \;\leftarrow\; \left(\frac{1}{N}\sum_{j=1}^N (\boldsymbol{\alpha}_k)_j\right)\mathbf{1}_N,
\end{equation}
\emph{inside} the unrolled inner problem, so outer meta-gradients see the reduction throughout training. This is \emph{not} ``freeze an element-wise checkpoint and insert (\ref{eq:lr_scalarize}) only at evaluation'': the scalarized row in Table~\ref{ta:lr_study} comes from its own AMASS (or quadratic) meta-training run with the flag enabled end-to-end.

Table~\ref{ta:lr_study} compares these recipes under matched settings.
Each row is a separately meta-trained checkpoint; scalarized runs apply (\ref{eq:lr_scalarize}) throughout meta-training and evaluation.
On quadratics we use unrolling depth $16$, secant weight $\lambda_{\text{sec}}=1$, fixed rate scaling $(\gamma_1,\gamma_2)=(0.4,10^{-3})$, and buffer size $L=8$, with four seeds on the $N{=}2$ validation set and $N{=}10$ test set; entries report mean inner loss over the first $50$ unrolled steps (mean$\pm$std).
On 3DPW we evaluate two AMASS-trained checkpoints with buffer $L=4$ and $\lambda_{\text{sec}}=1$, following the same protocol as Table~\ref{ta:hmr_results}: best PA-MPJPE (mm) over $50$ inner steps.
HMR serves here as a difficult, highly non-convex downstream example in contrast to the controlled quadratics: on the quadratic surrogate, scalarized meta-training attains lower mean inner loss, whereas on 3DPW element-wise meta-training achieves better PA-MPJPE by about $0.22\,\mathrm{mm}$.
We read this as evidence that difficult, structured objectives benefit from coordinate-wise learning-rate adaptation, while the simple quadratic meta-task favors a uniform step.
This coordinate-wise parameterization is consistent with prior learned optimizers~\citep{gartner2023transformer,metz2022practical}.

\begin{table}[t]
\centering
\caption{\textbf{Element-wise versus scalarized meta-training of $\boldsymbol{\alpha}_k$.}}
\label{ta:lr_study}

\small
\begin{tabular}{l|c|c|c}
\toprule

\multirow{2}{*}{\textbf{Training recipe}} &
\multicolumn{2}{c|}{\textbf{Quadratic}} &
\multirow{2}{*}{\textbf{3DPW}} \\

& \textbf{$N{=}2$} &
\textbf{$N{=}10$} & \\

\midrule

Scalar
& $-2.27 \pm 0.19$
& $-0.86 \pm 0.31$
& $51.80$ \\

Vector
& $3.07 \pm 0.21$
& $7.33 \pm 0.30$
& $51.58$ \\

\bottomrule
\end{tabular}
\end{table}

\subsection{Inputs and Hidden Dimension Ablations} \label{sec:ablations}
\begin{table}[h!]
  \centering
  \caption{\textbf{Varying hidden dimension and encoder inputs.} Quadratic validation set errors for different configurations of the hidden dimension and selected inputs to the encoder.}
  \label{tab:ablations}
  \begin{tabular}{c|ccccc|c}
    \toprule
    \textbf{Hidden dim.} & \multicolumn{5}{c|}{\textbf{Encoder Inputs}} & \\
    $d_{\text{hidden}}$ & $\bb{x}_{k-1}$ & $\bb{p}_{k-1}$ & $\bb{d}_{k-1}$ & $\bb{g}_{k-1}$ & $\bb{q}_{k-1}$ & Valid. loss \\
    \midrule
    128 & \cmark & \xmark & \xmark & \cmark & \xmark & -2.09 
    \\
    128 & \cmark & \xmark & \cmark & \cmark & \xmark & -2.16 
    \\
    128 & \cmark & \cmark & \xmark & \cmark & \cmark & -2.51 
    \\
    64  & \cmark & \cmark & \cmark & \cmark & \cmark & -1.90 
    \\
    \midrule
    \textbf{128} & \cmark & \cmark & \cmark & \cmark & \cmark & \textbf{-2.56} 
    \\
    \bottomrule
  \end{tabular}
\end{table}

We conducted a series of ablation studies on our model components. All experiments were carried out on the validation set, which comprises 32 quadratic functions as detailed in Appendix~\ref{sec:app_quad}. Table~\ref{tab:ablations} summarizes the results. We evaluated the impact of varying the hidden dimension as well as different combinations of inputs to the encoder. The selected configuration is highlighted in bold.


\subsection{Training with Limited Data} \label{app:exps_limited}
\begin{table}[t]
\centering
\scriptsize
\caption{\textbf{Data efficiency study.} 
L-SR1 trained on fractions of AMASS and evaluated on 3DPW.}
\label{ta:data_fractions}
\begin{tabular}{lcccc}
\toprule
\textbf{Fraction of Data}   & 80\%    & 10\%  & 1\%   & 0.1\% \\
\midrule
\textbf{PA-MPJPE}           & 51.67   & 53.03 & 56.14 & 70.78 \\
\bottomrule
\end{tabular}
\end{table}

To assess generalization beyond the training data, we trained L-SR1 on fractions of the AMASS dataset (80\%, 10\%, 1\% and 0.1\%) and evaluated on the full 3DPW test set. Even when trained on only 10\% of the data, L-SR1 achieved a test error of 53.03, which still outperforms LGD trained on the full dataset (55.90). With only 1\% of the data, the error is 56.14. Results are summarized in Table~\ref{ta:data_fractions}.

\newpage
\section{Implementation Details} \label{ap:implementations}

\subsection{L-SR1 Modules Architectures}
\label{ap:lsr1_arch}
Our proposed L-SR1 model comprises three learnable modules: an Input Encoder $\mathcal{E}$, a Vector Generator $\mathcal{P}$, and a Learning Rate (LR) Generator $\mathcal{G}$. 

All modules operate element-wise and share a common MLP-based architecture, detailed in Tables~\ref{ta:mlp} and~\ref{ta:basic}.
Concretely, inputs to each module are tensors of shape $B \times N \times d_{\text{in}}$, where $B$ denotes the batch size and $N$ the problem dimensionality.
The MLP is applied independently to each coordinate: all linear layers act exclusively on the feature dimension $d_{\text{in}}$, with parameters shared across the $N$ dimensions.
As a result, no mixing across coordinates occurs within the MLP itself.

Batch normalization is applied along the feature dimension by temporarily permuting the input to shape $B \times d_{\text{in}} \times N$, applying $\texttt{BatchNorm1d}(d_{\text{in}})$, and permuting the tensor back.
This ensures consistent normalization across batch elements and coordinates while preserving the element-wise structure of the computation.

We set $d_{\text{hidden}} = 128$ in all experiments. As described in Section~4.1 and justified in Appendix ~\ref{sec:ablations}, the Input Encoder uses $d_{\text{in}} = 5$ and $d_{\text{out}} = d_{\text{hidden}}$, while both the Vector and LR Generators use $d_{\text{in}} = d_{\text{hidden}}$ and $d_{\text{out}} = 1$.

\begin{table}[h]
\centering
\caption{\textbf{MLP architecture}}
\begin{tabular}{lll}
\toprule
\textbf{Layer} & \textbf{Type} & \textbf{Parameters} \\
\midrule
\texttt{fc1}   & Linear        & Input: $d_{\text{in}}$, Output: $d_{\text{hidden}}$ \\
\texttt{bn1}   & BatchNorm     & Features: $d_{\text{hidden}}$ \\
\texttt{prelu} & PReLU         & -- \\
\texttt{do1}   & Dropout       & -- \\
\texttt{MLP1}  & Basic Block   & Features: $d_{\text{hidden}}$ \\
\texttt{MLP2}  & Basic Block   & Features: $d_{\text{hidden}}$ \\
\texttt{fc2}   & Linear        & Input: $d_{\text{hidden}}$, Output: $d_{\text{out}}$ \\
\bottomrule
\end{tabular}
\label{ta:mlp}
\end{table}

\begin{table}[h]
\centering
\caption{\textbf{Basic Block Architecture}}
\begin{tabular}{lll}
\toprule
\textbf{Layer} & \textbf{Type} & \textbf{Parameters} \\
\midrule
\texttt{fc1}   & Linear        & Input: $d_{\text{in}}$, Output: $d_{\text{hidden}}$ \\
\texttt{bn1}   & BatchNorm     & Features: $d_{\text{hidden}}$ \\
\texttt{prelu} & PReLU         & -- \\
\texttt{do1}   & Dropout       & -- \\
\texttt{fc2}   & Linear        & Input: $d_{\text{hidden}}$, Output: $d_{\text{hidden}}$ \\
\texttt{bn2}   & BatchNorm     & Features: $d_{\text{hidden}}$ \\
\texttt{do2}   & Dropout       & -- \\
\bottomrule
\end{tabular}
\label{ta:basic}
\end{table}

\subsection{Experimental setup}

\subsubsection{Quadratic Functions} \label{sec:app_quad}

\paragraph{Data}
We generate random quadratic functions of the form
\begin{equation} \label{eq:quad}
    f(\mathbf{x}) = \frac{1}{2} \mathbf{x}^\top \mathbf{H} \mathbf{x} + \mathbf{b}^\top \mathbf{x},
\end{equation}
where $\mathbf{H} \in \mathbb{R}^{N \times N}$ is a positive semi-definite matrix and $\mathbf{b} \in \mathbb{R}^N$ is a random vector.

To construct $\mathbf{H}$, we draw a matrix $\mathbf{A} \in \mathbb{R}^{N \times N}$ from a standard normal distribution and define
\begin{equation}
    \mathbf{H} = \mathbf{A}^\top \mathbf{A},
\end{equation}
ensuring positive semi-definiteness. We compute the condition number of each $\mathbf{H}$ and discard those exceeding 1000. This process is repeated until full validation and test batches are acquired. Each matrix is then normalized to have unit Frobenius norm.
Independently, $\mathbf{b}$ is sampled from a standard normal distribution and normalized to have unit Euclidean norm.

During training batch generation, we relax the condition number constraint and accept all generated matrices, regardless of their conditioning. We use $N=2$ for both training and validation, and $N=10$ for testing. Training batches consist of 128 samples, while validation and test sets each contain 32 samples.

\paragraph{Hyperparameters}
We use the AdamW optimizer~\citep{loshchilov2017decoupled} for meta-training, with a fixed learning rate of $10^{-4}$, momentum parameters $\beta_1 = 0.9$ and $\beta_2 = 0.999$, and a weight decay coefficient of $\lambda = 0.01$. Training is conducted for 10{,}000 meta-iterations.

We set the buffer size to \( L = 8 \) and the number of unrolled iterations to \( K = 16 \). The Learning Rate Generator is configured with scaling parameters \( \gamma_1 = 0.4 \) and \( \gamma = 0.001 \). A secant constraint is applied with a weighting factor of \( \lambda_{\text{sec}} = 100 \). The learning rates for non-trainable optimizers were selected through hyperparameter tuning.

\subsubsection{Performance Profiles}
\label{ap:profiles}
\paragraph{Protocol}
We use performance profiles~\citep{dolan2002benchmarking} to aggregate solver performance across a set of benchmark problems.
Let \( P \) denote the set of problems and \( S \) the set of solvers.
For each solver \( s \in S \) and problem \( p \in P \), we define the performance measure
\[
m_{p,s} = \frac{\| \hat{\mathbf{x}}_{p,s} - \mathbf{x}^\ast \|_2}{\| \mathbf{x}_w - \mathbf{x}^\ast \|_2},
\]
where \( \hat{\mathbf{x}}_{p,s} \) is the solution found by solver \( s \) on problem \( p \) after \( K \) steps, \( \mathbf{x}_w \) is the worst solution among solvers, and \( \mathbf{x}^\ast \) is the global optimum.
The performance ratio is
\[
r_{p,s} = \frac{m_{p,s}}{\min \left( m_{p,s} \ :\ s \in S \right)},
\]
with the best solver achieving \( r_{p,s}=1 \).
The performance profile of solver \( s \) is then
\begin{equation}
    \rho_s(\tau) = \frac{1}{|P|}\text{size}\left( \{ p \in P \ :\ r_{p,s} \leq \tau \} \right),
\end{equation}
which measures the fraction of problems on which solver \(s\) is within a factor \(\tau\) of the best.

\paragraph{Data}
The quadratic functions used in this experiment are a subset of those defined in Eq.~\ref{eq:quad}, where \( \mathbf{b} = \mathbf{0} \) and \( \mathbf{H} \) is constrained to be diagonal. Our objective set comprises four such quadratic functions with condition numbers 1, 100, 1000, and 10000, along with the Rosenbrock and Rastrigin functions~\citep{simulationlib}. Each function is evaluated at input dimensions \( N = 50, 100, 250, 500 \), and \( 1000 \), yielding a total of 30 distinct optimization problems.

The solver set includes our proposed L-SR1 optimizer, both with and without PGSM, along with three non-trainable baselines: L-BFGS~\citep{nocedal1980updating}, Adam~\citep{kingma2014adam}, and AdaHessian~\citep{yao2021adahessian}, totaling six solvers. The learning rates for non-trainable optimizers were selected through hyperparameter tuning. L-BFGS was used with default settings, without an explicit line search.

Each trainable optimizer is meta-trained on three distinct tasks: a randomly generated quadratic function (as defined in Eq.~\ref{eq:quad}), the Rosenbrock function, and the Rastrigin function. For each task, a validation set of 8 fixed initial points is created and remains unchanged throughout meta-training. During each meta-iteration, a new training batch of 8 initial points is generated. Both training and validation are performed with $N=100$.

\paragraph{Hyperparameters}

\begin{table}[h!]
  \caption{\textbf{Hyperparameters used in Performance profiles}}
  \label{ta:params}
  \centering
  \begin{tabular}{lcccccc}
    \toprule
    \textbf{Parameter} & \multicolumn{2}{c}{\textbf{Quadratics}} & \multicolumn{2}{c}{\textbf{Rosenbrock}} & \multicolumn{2}{c}{\textbf{Rastrigin}} \\
    \cmidrule(r){2-3} \cmidrule(r){4-5} \cmidrule(r){6-7}
                      & Train & Test & Train & Test & Train & Test \\
    \midrule
    LR gen. param. $\gamma_1$      & 0.1 & 0.1 & 0.1 & 0.1   & 0.1 & 0.1   \\
    LR gen. param. $\gamma_2$      & 0.001 & 0.001   & 0.001 & 0.001   & 0.001 & 0.001   \\
    Buffer size $L$                     & 16  & 64    & 32  & 64    & 32  & 64    \\
    Unrolled iterations $K$             & 32  & -     & 64  & -     & 64  & -     \\
    Secant loss weight $\lambda_{\text{sec}}$ & 10  & -     & 1   & -     & 1   & -     \\
    \bottomrule
  \end{tabular}
\end{table}

We use the AdamW optimizer~\citep{loshchilov2017decoupled} for meta-training, with a fixed learning rate of $10^{-4}$, momentum parameters $\beta_1 = 0.9$ and $\beta_2 = 0.999$, and a weight decay coefficient of $\lambda = 0.01$. Training is conducted for 10{,}000 meta-iterations. Used hyperparameters are summarized in Table~\ref{ta:params}.

\subsubsection{Monocular Human Mesh Recovery (HMR)}
\label{ap:hmr}
\paragraph{Data}
We follow the training and evaluation protocol of \citep{song2020human}. Training is performed on the AMASS dataset \citep{AMASS}\footnote{The AMASS \citep{AMASS} dataset an aggregation of the following datasets \citep{AMASS_ACCAD, AMASS_BMLhandball, AMASS_BMLmovi, AMASS_BMLrub, AMASS_CMU, AMASS_DanceDB, AMASS_DFaust, AMASS_EyesJapanDataset, AMASS_GRAB, AMASS_GRAB-2, AMASS_HDM05, AMASS_HUMAN4D, AMASS_HumanEva, AMASS_KIT-CNRS-EKUT-WEIZMANN, AMASS_KIT-CNRS-EKUT-WEIZMANN-2, AMASS_KIT-CNRS-EKUT-WEIZMANN-3, AMASS_MoSh, AMASS_MOYO, AMASS_PosePrior, AMASS_SFU, AMASS_SOMA, AMASS_TCDHands, AMASS_TotalCapture, AMASS_WheelPoser}.}, which comprises SMPL body models \citep{loper2023smpl} parameterized by $(\boldsymbol{\beta}_{\text{gt}}, \boldsymbol{\theta}_{\text{gt}})$. At each training iteration, a batch of 2D joints is generated by projecting the corresponding 3D joints onto a randomly sampled camera view.

\begin{figure}[t]
\centering
\begin{subfigure}[t]{0.48\textwidth}
    \centering
    \includegraphics[width=\textwidth]{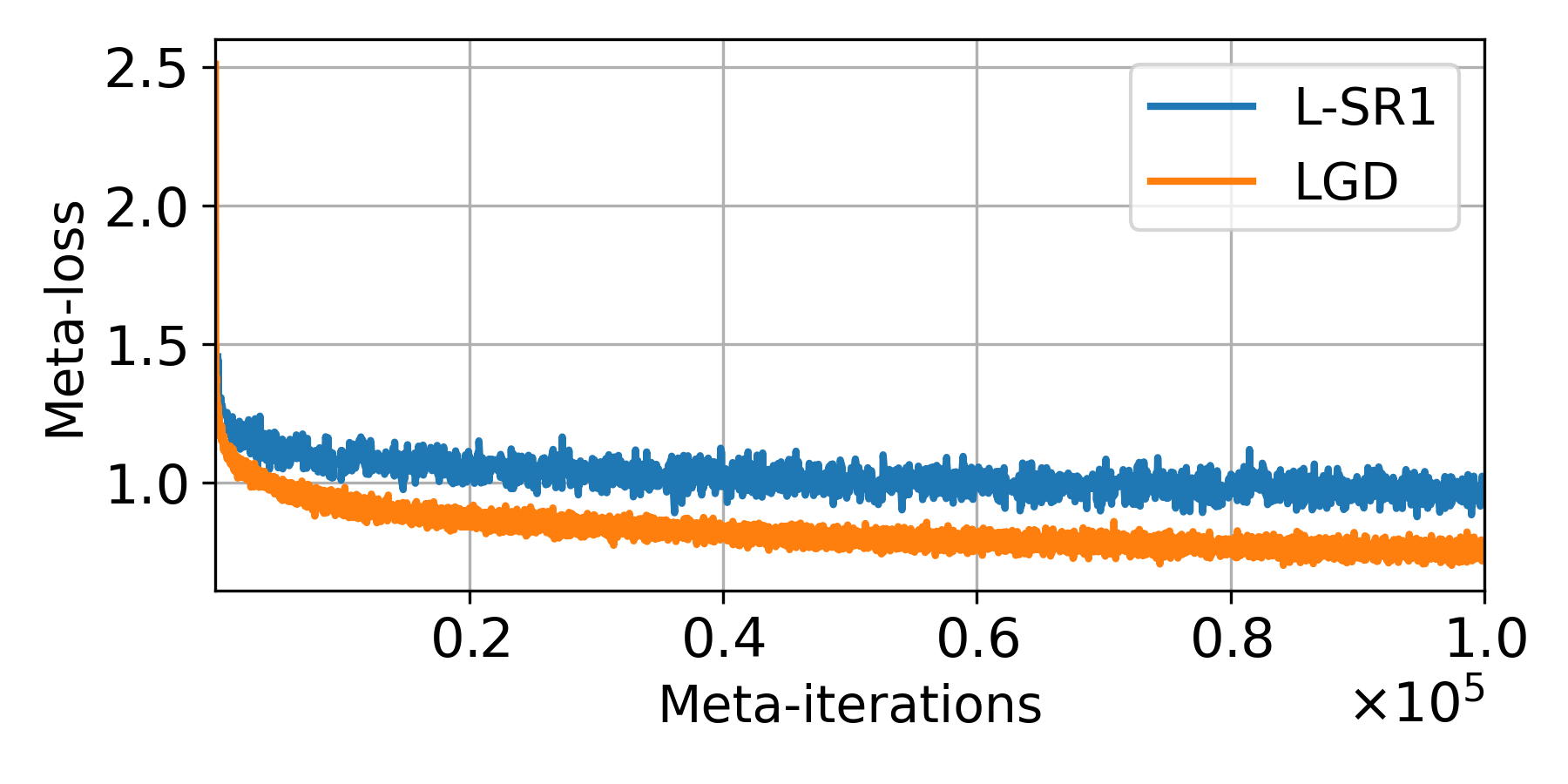}
    \caption{\textbf{HMR meta-loss on AMASS \citep{AMASS}.}}
    \label{fig:hmr_train}
\end{subfigure}
\hfill
\begin{subfigure}[t]{0.48\textwidth}
    \centering
    \includegraphics[width=\textwidth]{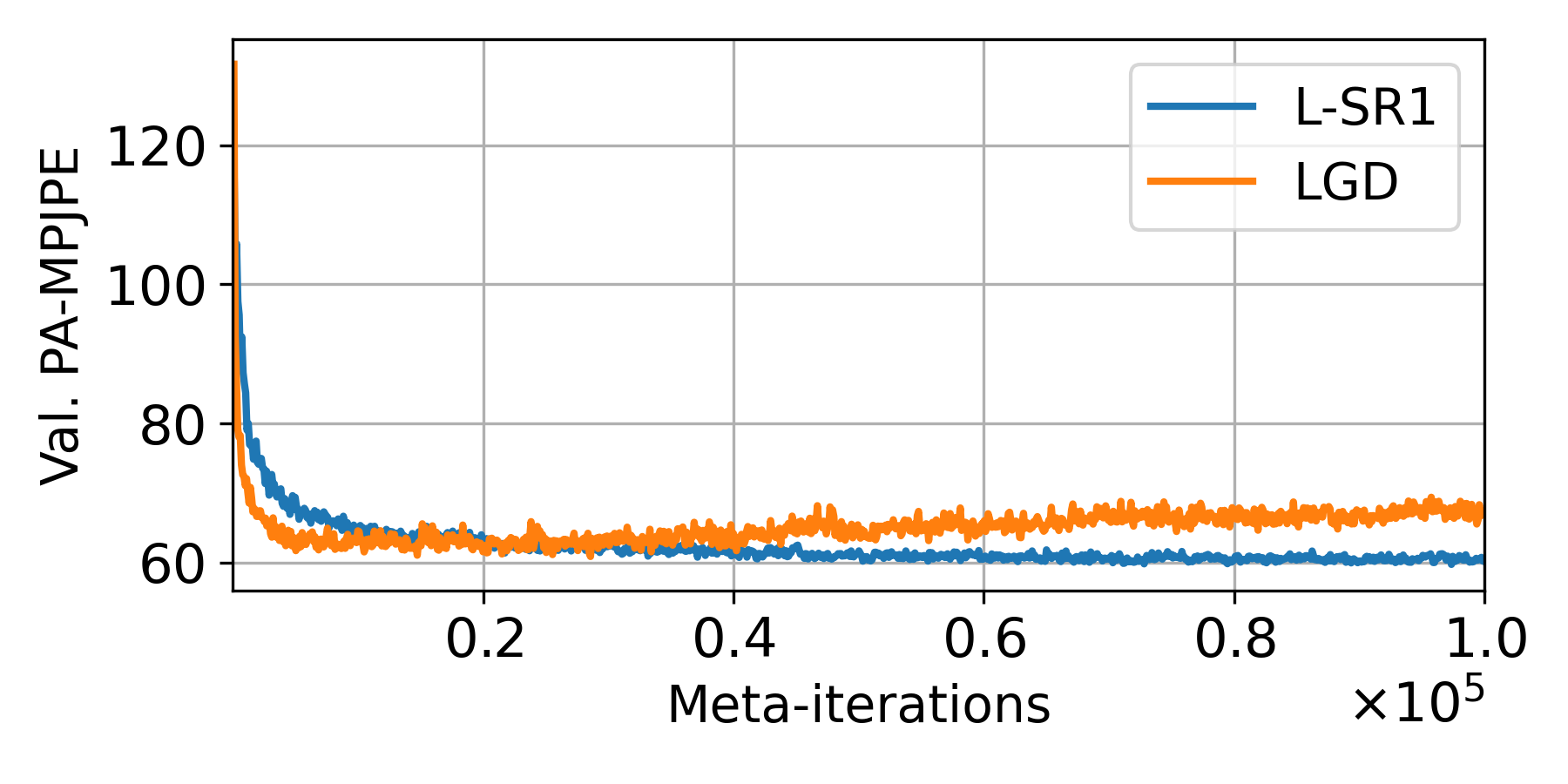}
    \caption{\textbf{HMR validation error on 3DPW \citep{vonMarcard2018}.}}
    \label{fig:hmr_valid}
\end{subfigure}
\caption{\textbf{HMR meta-training on AMASS and validation erros on 3DPW.} Shown first 100K iterations. Our meta-loss is higher as it has the secant loss added to it. }
\label{fig:hmr_meta_training}
\end{figure}

Validation and testing are conducted on the 3DPW dataset \citep{vonMarcard2018} using the same protocol as \citep{song2020human}, utilizing the provided 2D joint detections obtained via OpenPose \citep{cao2019openpose}. During training, we evaluate our model on the official 3DPW validation set and retain the checkpoint that achieves the best validation performance. The results reported in the paper are obtained by evaluating this best-performing model on the official 3DPW test set. Meta-training and validation graphs are given in Fig.~\ref{fig:hmr_meta_training}.

\paragraph{Hyperparameters}
We meta-train for 400K iterations using the AdamW optimizer~\citep{loshchilov2017decoupled}, with an initial learning rate of $10^{-3}$, which is decayed by a factor of $\gamma = 0.8$ every 20K iterations. The momentum parameters are set to $\beta_1 = 0.9$ and $\beta_2 = 0.999$, and the weight decay coefficient is $\lambda = 0.01$.

We configure the meta-optimization process with a buffer size of \( L = 4 \) and \( K = 8 \) unrolled iterations. The Learning Rate Generator is parameterized with scaling factors \( \gamma_1 = 0.1 \) and \( \gamma = 0.001 \). The secant constraint is weighted with \( \lambda_{\text{sec}} = 1 \).

\section{Limitations} \label{sec:limits}
Our evaluation is intentionally focused: controlled quadratics and performance profiles on standard analytic objectives, plus HMR as a challenging non-convex downstream task.
We do not claim universal superiority over every optimizer on every problem class within this scope.

L-SR1 carries higher per-iteration cost than first-order methods such as Adam~\citep{kingma2014adam} (Appendix~\ref{sec:compute}).
Reported HMR timings count full inner iterations, including SMPL forward kinematics and rendering; fewer iterations can still yield favorable wall-clock when the inner objective dominates, but the trade-off is task-dependent.

The learned modules are meta-trained on specific task distributions (random quadratics and, for HMR, AMASS).
Transfer to markedly different objectives may require re-training.
Finally, coordinate-wise learning rates and limited-memory curvature improve difficult problems in our study, but they add hyperparameters (buffer size, unrolling depth, and $\lambda_{\text{sec}}$) that must be set for new settings.
\newpage
\section{HMR Qualitative Results}
\label{ap:hmr_results}
\begin{figure}[h!]
\centering
\includegraphics[width=0.9\textwidth]{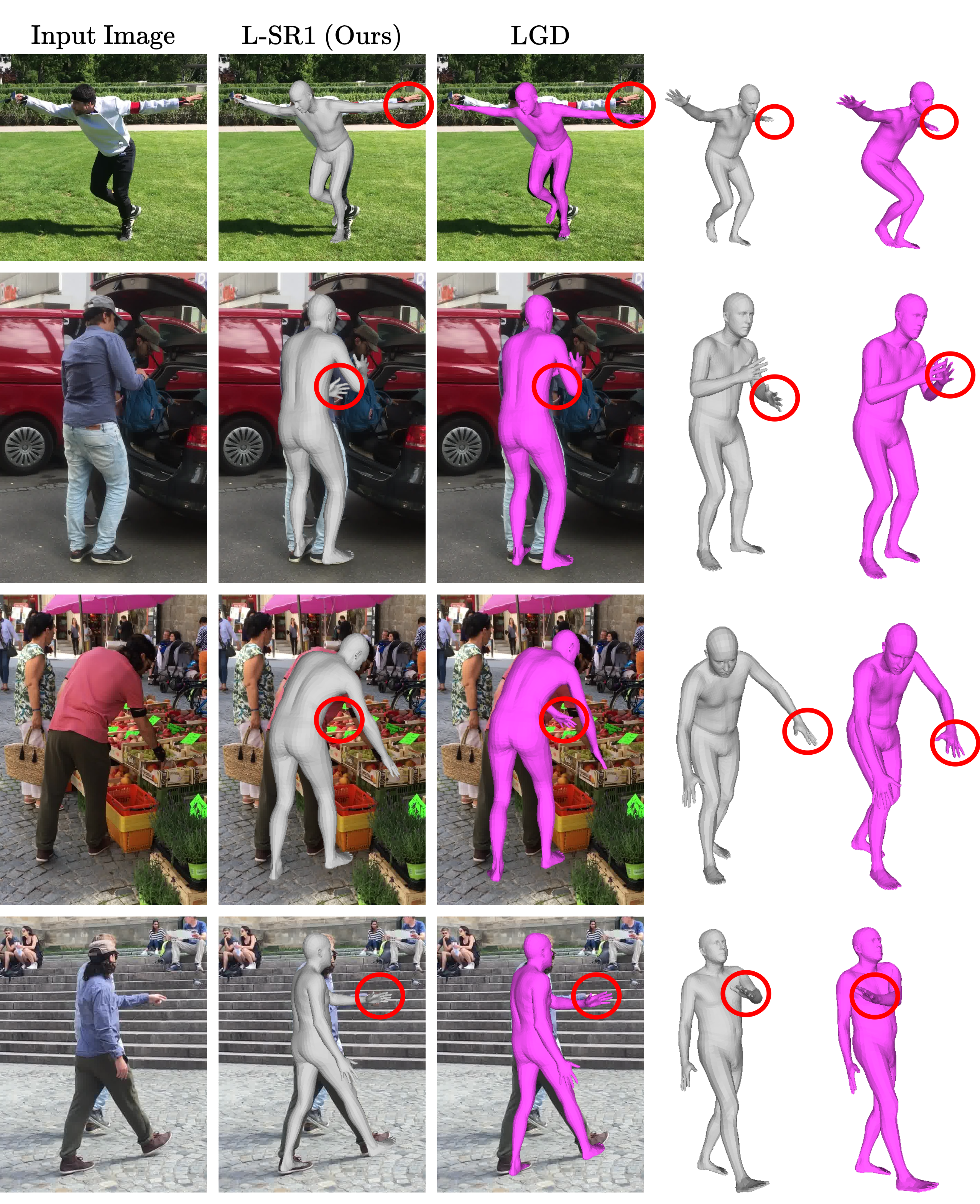}
\caption{
\textbf{Qualitative comparison of L-SR1 (ours) and LGD HMR results.}
Meshes optimized with L-SR1 (Ours) are shown in white, while those from LGD~\citep{song2020human} are shown in pink. Regions of interest are highlighted with red circles.
}
\label{fig:sup_hmr}
\end{figure}

\newpage
\section{Theoretical Background} \label{app:bg}

\begin{algorithm}[h]
\small
\caption{Quasi-Newton (QN) Optimization}\label{alg:qn}
\begin{algorithmic}
\REQUIRE Objective function $f \in \Ccal^2$, initial point $\mathbf{x}_0 \in \RR^n$
\STATE Initialize $\mathbf{B}_0 \gets \mathbf{I}$
\FOR{$k = 1,2,\ldots$} 
    \STATE $\mathbf{d}_k \gets -\mathbf{B}_{k-1} \nabla f(\mathbf{x}_{k-1})$
    \COMMENT{Compute descent direction}
    
    \STATE Choose $\alpha_k$ such that
    $f(\mathbf{x}_{k-1} + \alpha_k \mathbf{d}_k) < f(\mathbf{x}_{k-1})$
    \COMMENT{Find step size}
    
    \STATE $\mathbf{x}_k \gets \mathbf{x}_{k-1} + \alpha_k \mathbf{d}_k$
    \COMMENT{Optimization step}
    
    \STATE $\mathbf{B}_k \gets
    \textsc{Update}\!\left(
        \mathbf{B}_{k-1},\,
        \mathbf{x}_k - \mathbf{x}_{k-1},\,
        \nabla f(\mathbf{x}_k) - \nabla f(\mathbf{x}_{k-1})
    \right)$
    \COMMENT{Update $\mathbf{B}_k$}
\ENDFOR
\ENSURE $\mathbf{x}^* \gets \mathbf{x}_k$
\end{algorithmic}
\end{algorithm}

\begin{algorithm}[h]
\small
\caption{SR1 Update of Inverse Hessian Estimate}\label{alg:sr1}
\begin{algorithmic}
\REQUIRE Inverse Hessian estimate $\mathbf{B}_{k-1}$,
         step $\mathbf{p}_k \triangleq \mathbf{x}_{k} - \mathbf{x}_{k-1}$,
         gradient difference $\mathbf{q}_k \triangleq \mathbf{g}_{k} - \mathbf{g}_{k-1}$
         
\STATE $\mathbf{v} \gets \mathbf{p}_k - \mathbf{B}_{k-1} \mathbf{q}_k$

\IF{$\mathbf{v} \not\perp \mathbf{q}_k$}
    \STATE $\mathbf{B}_k \gets \mathbf{B}_{k-1}
    + \dfrac{\mathbf{v}\mathbf{v}^{\top}}{\mathbf{v}^{\top}\mathbf{q}_k}$
    \COMMENT{SR1 update}
\ELSE
    \STATE $\mathbf{B}_k \gets \mathbf{B}_{k-1}$
    \COMMENT{Skip update}
\ENDIF

\ENSURE $\mathbf{B}_k$
\end{algorithmic}
\end{algorithm}

\subsection{Quasi-Newton Approach}

Consider the Newton Method (NM), which utilizes the full Hessian matrix for preconditioning via directional adjustments. The NM minimization direction is defined as 
\begin{equation}
    \bb{d}_\text{NM} = - \bb{H}^{-1}(\bb{x}_k) \nabla f (\bb{x}_k),
\end{equation}
where $\bb{H}^{-1}(\bb{x}_k)$ represents the inverse Hessian matrix evaluated at $\bb{x}_k$. Notably, when $f$ is quadratic, NM theoretically converges in a single iteration. However, the practicality of NM is often hindered by the challenge of computing and inverting the Hessian matrix. Consequently, a class of second-order optimization methods, termed Quasi-Newton (QN), emerged, aiming to approximate the inverse Hessian matrix, denoted $\bb{B}$, during the optimization process. This involves updating $\bb{B}$ iteratively alongside each optimization step. Algorithm \ref{alg:qn} presents a summary of the QN optimization approach.

\subsection{Symmetric-Rank-One (SR1)}

One prominent Quasi-Newton (QN) method is the Symmetric-Rank-One (SR1) technique, which involves iteratively accumulating symmetric rank-one matrices to estimate $\bb{B}$. Let $\bb{g}_k = \nabla f(\bb{x}_k)$ and $\bb{B}_k = \bb{H}^{-1}(\bb{x}_k)$ denote the gradient and inverse Hessian at step $k$, respectively. Considering the linear approximation of $\bb{g}_{k}$ as:

\begin{equation} \label{eq:approx}
    \bb{g}_{k} = \bb{g}_{k-1} + \bb{H}(\bb{x}_{k-1})\left( \bb{x}_{k} - \bb{x}_{k-1} \right),
\end{equation} 

and defining $\bb{p}_k = \bb{x}_{k} - \bb{x}_{k-1}$ and $\bb{q}_k = \bb{g}_{k} - \bb{g}_{k-1}$, equation (\ref{eq:approx}) reduces to:

\begin{equation} \label{eq:secant}
    \bb{p}_k =  \bb{B}_{k} \cdot \bb{q}_k.
\end{equation}

This equation imposes a constraint directly on $\bb{B}_{k}$, known as the "secant constraint". 

The SR1 method involves updating $\bb{B}$ with rank-one matrices of the form:

\begin{equation} \label{eq:sr1_upd}
     \bb{B}_k \gets \bb{B}_{k-1} + \bb{u}\bb{v}^{\top},
\end{equation}

where $\bb{u},\bb{v} \in \RR^N$. Assuming $\bb{B}_{k-1}$ is symmetric, $\bb{B}_k$ is symmetric as well. Enforcing the secant constraint given in equation (\ref{eq:secant}) on $\bb{B}_k$ yields:

\begin{equation}
        \bb{p}_k =  \left( \bb{B}_{k-1} + \bb{u}\bb{v}^{\top} \right) \cdot \bb{q}_k,
\end{equation}

which can be rearranged as:

\begin{equation}
    \bb{u} = \frac{\bb{p}_k - \bb{B}_{k-1} \bb{q}_k}{\bb{v}^{\top} \bb{q}_k}.
\end{equation}

This implies that for every $\bb{v}\not\perp \bb{q}_k$, a suitable $\bb{u}$ satisfying the secant constraint can be found. A common choice is $\bb{v} = \bb{p}_k - \bb{B}_{k-1} \bb{q}_k$, leading to the following SR1 update step:

\begin{equation} \label{eq:sr1}
    \bb{B}_k \gets \bb{B}_{k-1} + \frac{\left(\bb{p}_k - \bb{B}_{k-1} \bb{q}_k\right)\left(\bb{p}_k - \bb{B}_{k-1} \bb{q}_k\right)^{\top}}{\left( \bb{p}_k - \bb{B}_{k-1} \bb{q}_k \right)^{\top} \bb{q}_k}.
\end{equation}

The SR1 'Update B' process, summarized in Algorithm \ref{alg:sr1}, notably ensures the symmetry of the estimated $\bb{B}$, a desirable feature. However, it falls short in guaranteeing positivity, a crucially property, as emphasized in the subsequent lemma.

\begin{lemma}
Let $f \in \Ccal^2$ be an objective function, and $\bb{d}_\text{NM} = -\bb{B}\ \bb{g}$ represent the Newton direction. Then, $\bb{B} \succcurlyeq \bb{0}$ ensures that $\bb{d}_\text{NM}$ is a descent direction.
\end{lemma}

\begin{proof}
    $\bb{d}_\text{NM}$ is a descent direction if and only if the directional derivative of $f$ in the direction $\bb{d}_\text{NM}$ satisfies $f'_{\bb{d}_\text{NM}} \leq 0$. Since $f'_{\bb{d}} = \bb{g}^{\top} \bb{d}$,
    \begin{equation}
        0 \geq f'_{\bb{d}_\text{NM}} = \bb{g}^{\top}\bb{d}_\text{NM} = -\bb{g}^{\top}\bb{B}\ \bb{g} \Leftrightarrow \bb{g}^{\top}\bb{B}\ \bb{g} \geq 0
    \end{equation}
    Hence, $\bb{B} \succcurlyeq \bb{0}$ is a sufficient condition ensuring $\bb{d}_\text{NM}$ is a descent direction.
\end{proof}

Consequently, several methods have been proposed to either eliminate 'bad' directions or ensure positive matrices through more sophisticated schemes.


\end{document}